\documentclass[10pt,twocolumn,letterpaper]{article}

\usepackage{cvpr}              

\usepackage{dsfont}
\usepackage{algorithm}
\usepackage{algpseudocode}
\usepackage{wrapfig}

\definecolor{cvprblue}{rgb}{0.21,0.49,0.74}
\usepackage[pagebackref,breaklinks,colorlinks,allcolors=cvprblue]{hyperref}

\def\paperID{8157}
\def\confName{CVPR}
\def\confYear{2026}

\newcommand{\mytitle}{Computer Vision with a Superpixelation Camera}

\title{\mytitle}

\author{Sasidharan Mahalingam\\
Portland State University\\
{\tt\small samahali@pdx.edu}
\and
Rachel Brown$^{*}$\\
Willamette University\\
{\tt\small rabrown3@willamette.edu}
\and
Atul Ingle$^{*}$ \\
Portland State University\\
{\tt\small atul.ingle@pdx.edu}
}

\begin{document}

\twocolumn[{%
\renewcommand\twocolumn[1][]{#1}%
\vspace{-0.2in}
\maketitle

\begin{center}
    \centering
    \captionsetup{type=figure}
    \vspace{-0.35in}
    \includegraphics[width=\textwidth]{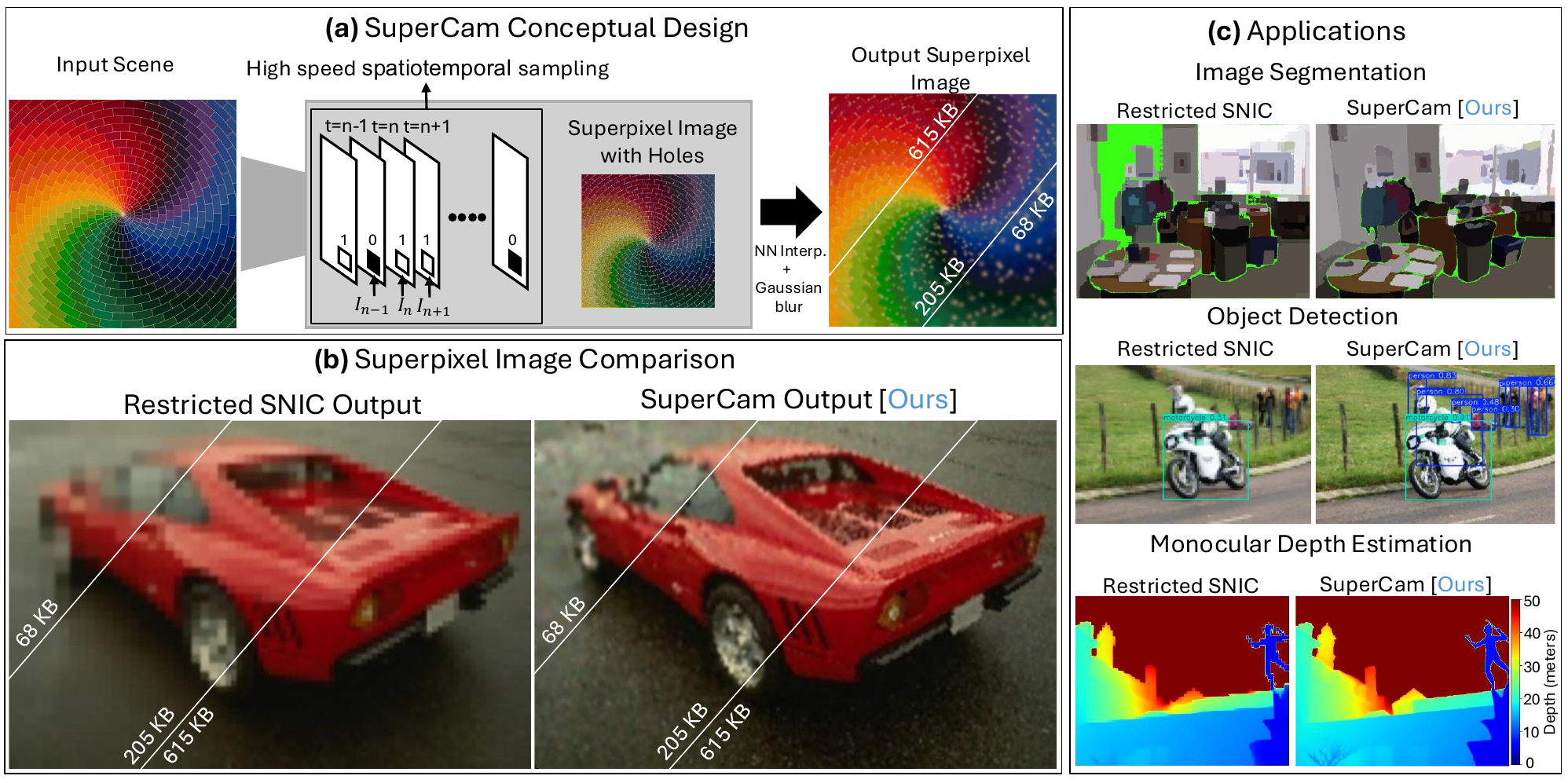}
    \captionof{figure}{\textbf{Summary of SuperCam design, output, and applications.} We propose a new camera design that generates superpixel images on the fly without storing/capturing the full input image.
    \textbf{(a) SuperCam Conceptual Design.} The proposed design, which is implemented as a passive Single Photon Avalanche Diode (SPAD) Sensor.
    We adaptively sample the exposed pixels to generate a sparse superpixel image.
    This processing happens on-sensor.
    This sparse data is read out and filled off-sensor using nearest-neighbor interpolation and Gaussian blur to generate the superpixel image.
    \textbf{(b) Superpixel Image Comparison.} As SuperCam generates the superpixel image from sparse data on the fly, it performs better than the conventional superpixel algorithms under resource constrained scenarios.
    Shown are sample images with different memory settings; 68KB, 205KB and 615KB. 
    \textbf{(c) Applications.} Three computer vision applications comparing SuperCam with a memory-constrained implementation of SNIC (Image Segmentation using SAMv2, Object Detection using YOLOv12 and Monocular Depth Estimation using DepthAnythingv2).
    \label{fig:teaser}
    }
\end{center}%
}]
\makeatletter{\renewcommand*{\@makefnmark}{}
\footnotetext{
$^*$ Equal contribution.

This material is based upon work supported by NSF ECCS-2138471; and the U.S. Department of Energy, Office of Fusion Energy Sciences, under Award Number DE-SC0025802.
}\makeatother}
\begin{abstract}
Conventional cameras generate a lot of data that can be challenging to process in resource-constrained applications. Usually, cameras generate data streams on the order of the number of pixels in the image. However, most of this captured data is redundant for many downstream computer vision algorithms. We propose a novel camera design, which we call SuperCam, that adaptively processes captured data by performing superpixel segmentation on the fly. We show that SuperCam performs better than current state-of-the-art superpixel algorithms under memory-constrained situations. We also compare how well SuperCam performs when the compressed data is used for downstream computer vision tasks. Our results demonstrate that the proposed design provides superior output for image segmentation, object detection, and monocular depth estimation in situations where the available memory on the camera is limited. We posit that superpixel segmentation will play a crucial role as more computer vision inference models are deployed in edge devices. SuperCam would allow computer vision engineers to design more efficient systems for these applications.
\vspace{-4.5em} 
\end{abstract}
\section{Introduction}
\label{sec:intro}
The notion of an image as a collection of square shaped pixels arranged in a uniform grid is a \emph{fait accompli} that both classical and modern (deep-learning-based) computer vision system are built on.
In this paper we ask: what if the raw ``images'' coming off an image sensor were instead loosely-grouped clusters with arbitrary shapes adapted to the scene being imaged?
 We call this hypothetical image sensor a ``superpixelation camera'' or SuperCam for short.

There is a long line of work on representing images as superpixels, groups of pixels that are usually clustered based on pixel intensity or other nuanced measures of pixel similarity. 
These superpixel segmentation algorithms start with the original high resolution image and generate a superpixel simplification representation from that high resolution image. 
A SuperCam is quite different; it is not merely a postprocessing algorithm that creates a parsimonious image representation from a high resolution image.
We envision a sensor that avoids capturing the high resolution image altogether and directly outputs superpixels (Fig.~\ref{fig:teaser}(a-b)).

The motivation for this is partly academic and partly driven by the need for more resource-efficient image sensors.
A superpixel approach more closely mimics how biological vision systems work---the human eye, for example, does not really form a high resolution image like a camera. Edge detection and grouping of perceptually similar scene regions happens as early as the retina itself, quite early in the visual processing chain. By contrast, image sensors contain hundreds of millions of pixels packed into a tiny area, consuming power and requiring large bandwidth only for most of it to be simply thrown away later\footnote{Interestingly, the inventor of the pixel lamented his decision to make them square-shaped \cite{wiredarticle}.
}.

We do not and cannot aim for raw image quality with SuperCam, so we eschew image quality metrics in favor of results showing direct applications.
Our results demonstrate that with SuperCam images, it is still possible to perform many computer vision tasks, including semantic segmentation, object detection, and monocular depth estimation, with a significant reduction in memory requirements over using a conventional image (Fig.~\ref{fig:teaser}(c)).
Our instantiation of SuperCam also comes with a ``tuning knob'' that can be adjusted to smoothly trade-off fidelity for memory requirements in these computer vision tasks as needed.
We show results at different ``memory settings'' and show a graceful increase in performance (measured by the task-specific quality metrics) as the number of superpixels, and the on-sensor memory needed to store them, is increased.

\vspace{0.5em}
\section{Related Work}
 \label{sec:background}
Because SuperCam is a novel superpixel algorithm built atop an unconventional camera using a sparse data stream, our related work explores both existing superpixel algorithms, methods for compressive sensing, and other unconventional vision sensors.

\smallskip
\noindent\textbf{Superpixel algorithms.} 
Superpixel segmentation, first introduced by Ren and Malik \cite{superpixel-survey1}, refers to segmenting an image into smaller regions or clusters that are locally similar while preserving the overall coherence of the image.
We refer the reader to \cite{superpixel-survey1, superpixel-survey2} for more detailed surveys of the field.
Superpixel images can be very useful for computer vision tasks because they reduce the complexity of the input, simplifying the search space. 
Many superpixel algorithms have been proposed to increase efficiency and improve output quality for various computer vision tasks: semantic segmentation \cite{sem_seg_application1, sem_seg_application2, sem_seg_application3, sem_seg_application4, sem_seg_application5}, 3D reconstruction \cite{superpixel-survey1, 3d_recon_application1, 3d_recon_application2, 3d_recon_application3, 3d_recon_application4}, object detection \cite{superpixel-survey1, obj_det_application1, obj_det_application2, obj_det_application3, obj_det_application4}, depth estimation \cite{superpixel-survey1, depth_application1, depth_application2, depth_application3, depth_application4}, tracking \cite{superpixel-survey1, obj_track_application1, obj_track_application2, obj_track_application3, obj_track_application4} and optical flow \cite{superpixel-survey1, opt_flow_application1, opt_flow_application2, opt_flow_application3, opt_flow_application4}. 

Although there have been many recent technological advances in superpixel segmentation, many of these techniques have significant memory requirements that prohibit their use in lightweight edge-computing applications.
This includes algorithms that rely on learned features \cite{DAL-HERS,semasuperpixel}, unsupervised end-to-end networks \cite{LNSNET, SEEK}, and transformer-based solutions \cite{CLUSTSEG}.
There are also a number of superpixel algorithms which can be light-weight but require the entire input image as input, such as graph-based \cite{graph_based_sp1, ERS}, clustering-based \cite{SLIC, SNIC} and energy-based \cite{TurboPixels, CRS, SEEDS} algorithms, as well as recent Bayesian techniques \cite{bayesian_superpixels, BASS}. 
By contrast, we do not have access to the entire image and would instead like to create a sensor whose raw data output is the superpixel information.

We therefore consider clustering-based algorithms such as SNIC \cite{SNIC} and SLIC \cite{SLIC} to be the most comparable methods to our work. In SNIC \cite{SNIC}, segmentation is performed by adding and removing pixel elements from a queue until all pixels are labeled. The feature space used to create the queue is derived from a weighted combination of pixel distances and intensities. Similarly, SLIC \cite{SLIC} clusters pixels in a combined five-dimensional color and image plane space to efficiently generate compact, nearly uniform superpixels. Of these methods, we compare against SNIC \cite{SNIC} as it is widely used, provides the lowest error metrics among the non-learning based methods \cite{AINET, SSN, semasuperpixel, LNSNET}, and meets all the necessary criteria for our input data stream.

\smallskip
\noindent
\textbf{Single-pixel and compressive image sensing.}
There is a rich line of work on single-pixel image sensing \cite{gibson2020single} starting with seminal work on compressed-sensing-based single-pixel camera that captures random linear projections and still recovers the original image \cite{duarte2008single}.
Our motivation with SuperCam is related in the sense that we aim to capture parsimonious representations of the original scene.
However, our end goal is different because we focus on performance metrics that have less to do with the image reconstruction quality, but more to do with the downstream computer vision tasks.

\smallskip
\noindent\textbf{Unconventional vision sensors.}
Image sensors with in-pixel and on-sensor compute capabilities can be used to implement intelligent vision sensing on edge devices.
Such compute capabilities have been demonstrated with both traditional CMOS-type pixels through ``pixel processor arrays'' for realtime feature tracking \cite{zhang2025focal, bose2025descriptor}, single-photon-sensitive pixel arrays with in-pixel compute \cite{ardelean2023computational}, and sensors with extremely low spatial resolution, coupled with a learning-based approach for specific vision tasks \cite{freeform_pixels}.
We envision future hardware implementations of SuperCam can leverage these advances in reconfigurable and reprogrammable image sensors.
Event sensors reduce data rate by only transmitting intensity changes \cite{gallego2020event}.
Our work is the spatial counterpart of event cameras---whereas event cameras transmit information when the \emph{temporal} change is large, SuperCam reduces the output data volume by only storing superpixel information for \emph{spatial} regions that are perceptually distinct in color and intensity information.
\section{Superpixelation Camera: Imaging Model}
\label{sec:ImagingModel}
In this section we present an abstract model for capturing scene information using a superpixelation camera.
Since such a camera does not yet exist, next we present a practical method for simulating (purely on a computer) and emulating (using existing camera hardware) one.
Instead of storing an image as a regular grid of pixel values, a SuperCam maintains an internal data structure of segment information in the form of segment boundaries and a single 8- or 24-bit intensity and color value associated with each segment.
Previous works \cite{CLUSTSEG, SNIC, SLIC, AINET, E2E-SIS, SSN} have used superpixels in the range of 1000 to 2000 for images from the BSD500\cite{BSD500} dataset, which are $321 \times 481$ in dimension. 
Unlike conventional images that require millions of pixels with 8- or 24-bit (intensity or color) information, the SuperCam data structure needs orders of magnitude lower memory.

Let $\mathcal{S}=(S_i, I_i)_{i=1}^N$ denote the set of $N$ segments maintained by the SuperCam's internal, on-sensor data structure.
Here $S_i$ denotes the boundary information and $I_i$ denotes the intensity information associated with the $i^\text{th}$ superpixel.
The image sensor acquires scene information by integrating photons (over a pre-specified exposure duration) at some arbitrarily chosen co-ordinates $(x,y)$ on the sensor plane, obtaining a measurement $\phi$ of the incident photon flux.
After each new measurement, the SuperCam data structure $\mathcal{S}$ is updated: all segments $S_i$ such that $(x,y) \in S_i$, must update their intensity estimates, $I_i$, based on the new measurement $\widehat{\phi}$.
This iterative update routine is shown in Fig.~\ref{alg:super-cam-alg}.

\begin{figure}
\begin{algorithm}[H]
    \caption{The SuperCam Algorithm}
     \begin{algorithmic}[1]
        \State $P \gets $ number of superpixels
        \State Divide the image into $P$ number of equal rectangles.
        \For {$i=1$ to $P$}
            \State \parbox[t]{200pt}{$(x_i,y_i) \gets $ randomly chosen sensor-plane position\strut}
            \State Expose $(x_i,y_i)$ for a fixed exposure time $\tau$.
            \State Calculate the intensity estimate $\widehat{\phi}(x_i,y_i)$
            \State \parbox[t]{200pt}{Initialize segments with co-ordinates $(x_i,y_i)$ and intensity estimates $\widehat{\phi}(x_i,y_i)$.\strut}
            \item[]
        \EndFor
        \State Fill the holes in the image using the intensity value of the nearest segment.
        \State Apply Gaussian blur.
    \end{algorithmic}
\end{algorithm}
\vspace{-0.2in}
\caption{\textbf{SuperCam pseudoscode.} SuperCam stores scene information as a collection of segment boundaries and their associated intensity values. Segment information is updated after a new measurement is available. We envision steps 4--7 of the algorithm to run on-sensor (in or near the camera pixels) and step 9--10 to be implemented off-sensor. The SuperCam algorithm is simple, does not require heavy compute and can be implemented with minimal hardware modifications. The nearest segment operation is parallelizable and can be implemented to run real-time off sensor.
\label{alg:super-cam-alg}}
\vspace{-0.2in}
\end{figure}

After all the initial superpixel boundaries and intensities are obtained, we fill in any holes in the image using nearest neighbor interpolation. 
We then apply a Gaussian blur kernel with a blur radius equal to the superpixel grid size in SuperCam.
We use separable blur kernels with different blur radii for superpixel grids that are rectangular in shape.
A derivation of the blur kernel size is given in the supplement.

\smallskip
\noindent
\textbf{SuperCam: Practical implementations.}
The abstract measurement model encompasses many existing image sensing techniques including a raster-scanning single-pixel sensor \cite{gibson2020single}, position-sensitive photomultiplier tubes \cite{kuroda1982position}, and even a conventional pixelated sensor with row/column addressable readout \cite{dierickx1996random}.
For the remainder of this paper we focus on one specific implementation of the SuperCam idea that relies on a single-photon sensitive camera sensors made up of a high-resolution array single-photon avalanche diode (SPAD) pixels with a single readout that reports the $(x,y)$ location of each new photon detection event.
SPAD cameras sense scene information at the finest spatio-temporal granularity possible, with on-sensor compute that can mimic a wide range of image sensing modalities including conventional CMOS/CCD pixels and event cameras \cite{sundar2023sodacam}.
Hence they provide a suitable testbed for implementing and testing the performance limits of SuperCam.

We simulate the SuperCam as SPAD array passively collecting photons.
We selectively expose the individual SPAD pixels to build the superpixel image in a single exposure.
As a proof of concept, we implement a passive SPAD simulator that generates individual photon arrivals over a single exposure and generates an intensity image using the quanta burst model proposed in \cite{quanta_burst}.
We also propose a passive SPAD simulator that simulates the given image with a mean photon per pixel setting that adaptively exposes the image to obtain an image that has minimum photon noise with a limited number of photon arrival simulation trials.

\subsection{SuperCam Emulation using a SPAD Camera}
Assuming a passive SPAD sensor array imaging a scene, the number of photons $Z(x,y)$ arriving at a sensor plane location $(x,y)$ can be modeled as a Poisson random variable \cite{poission_stat} 
    $
    P\{Z=k\} = \frac{(\phi \tau \eta)^ke^{-\phi \tau \eta}}{k!}
    $
where $\phi(x,y)$ is the photon flux (photons/second) arriving at the pixel $(x,y)$, $\tau$ is the exposure time, and $\eta$ is the quantum efficiency.
Each SPAD pixel can store at most 1 photon detection event, hence the pixel readout is a binary value $B(x,y)$ which follows a Bernoulli distribution: $P\{B=0\} = e^{-(\phi \tau \eta + r_q \tau)}$ and 
$P\{B=1\} = 1 - e^{-(\phi \tau \eta + r_q \tau)}$
where $r_q$ is the dark count rate, which refers to spurious counts that are not due to incident photons.

Starting from RGB intensity images from publicly available datasets, we simulate the binary streams of photon detections using the following procedure.
We choose a ``mean photons per pixel'' value of $p$, and adjust the exposure for each individual image to match this mean value.
Assuming that the incident photon flux $\phi$ is proportional to the value of the intensity image $I_i$, the probability that a SPAD pixel registers a 1 is given by, $P\{B=1\} = 1 - e^{-cI_i}$, where the value of $c$ is the exposure adjustment term.
Next, assuming there are $M$ pixels in the image and $N$ binary image frames captured by the camera, we select a value of $c$ to ensure that $\frac{1}{M} \sum_{i=1}^{M} (1 - e^{-cI_i}) = \frac{p}{N}$.
Assuming the incident photon flux is low enough that $cI_i \ll 1$, we apply a Taylor series approximation $1-e^{-cI_i} \approx cI_i$ which gives us a closed form expression for the per-image exposure scaling term: $c = \frac{p}{N I_\text{avg}}$
where $I_\text{avg}$ is the average of all the pixel values in the ground truth image.
Recovering the true pixel intensities from these binary-valued measurements involves summing the binary frames $S(x,y)$  and applying a log-compression \cite{quanta_burst}:
$
    \hat{\phi}(x,y) = -\ln(1 - S(x,y)/M) / (c \eta) - r_q/\eta.
$



\subsection{Comparisons}
Since there are no comparable baselines for cameras that generate superpixel images on the fly, we compare against modified versions of existing methods that have been restricted to use the same volume of on-sensor memory (subsequently referred to as ``Restricted''). Here we illustrate our method for matching the memory footprint using SNIC~\cite{SNIC}.

For an original input image with $M$ pixels and a target output number of $P$ superpixels, our (unoptimized implementation of the) SuperCam data structure which stores the superpixel boundaries and intensity values has a memory footprint $\sim$\emph{10P}. The off-the-shelf implementation of SNIC needs access to the raw RGB pixel values for the entire image so that they can be processed using a priority queue, which maintains a list of pixels most ``similar'' to the current pixel being processed. The memory footprint of SNIC therefore depends not only on $P$ but also on $M$ because the priority queue contains all pixel intensities. 
We produce a memory restricted implementation by first setting the memory budget (e.g., between 70-700KB), then adjusting the number of superpixels and image size such that the combined memory of the image and required data structure fit into the preset memory budget. 
In order to determine the ideal ratio between the number of superpixels and image size, we carried out coarsely sampled experiments for a range of settings and picked the value that produced the lowest Under Segmentation Error. For SNIC, the lowest error was obtained by allocating $5\times$ as much memory for the image data as for the number of superpixels.

Empirically, we found that applying Gaussian blur to the memory-constrained SNIC image using the same kernel size we applied to the SuperCam image improves the results of downstream computer applications for SNIC as well. We therefore include these blurred inputs for comparison, which have been labeled as ``SNIC with blur" in subsequent figures.


So far in this section we have described how we simulate a SuperCam from scene information captured at the granularity of individual photons.
In the next section, we will use simulated and emulated data to show SuperCam's capabilities for a wide range of computer vision tasks.


\begin{figure}[!ht]
    \centering
    \vspace{-0.1in}
    \includegraphics[width=\linewidth]{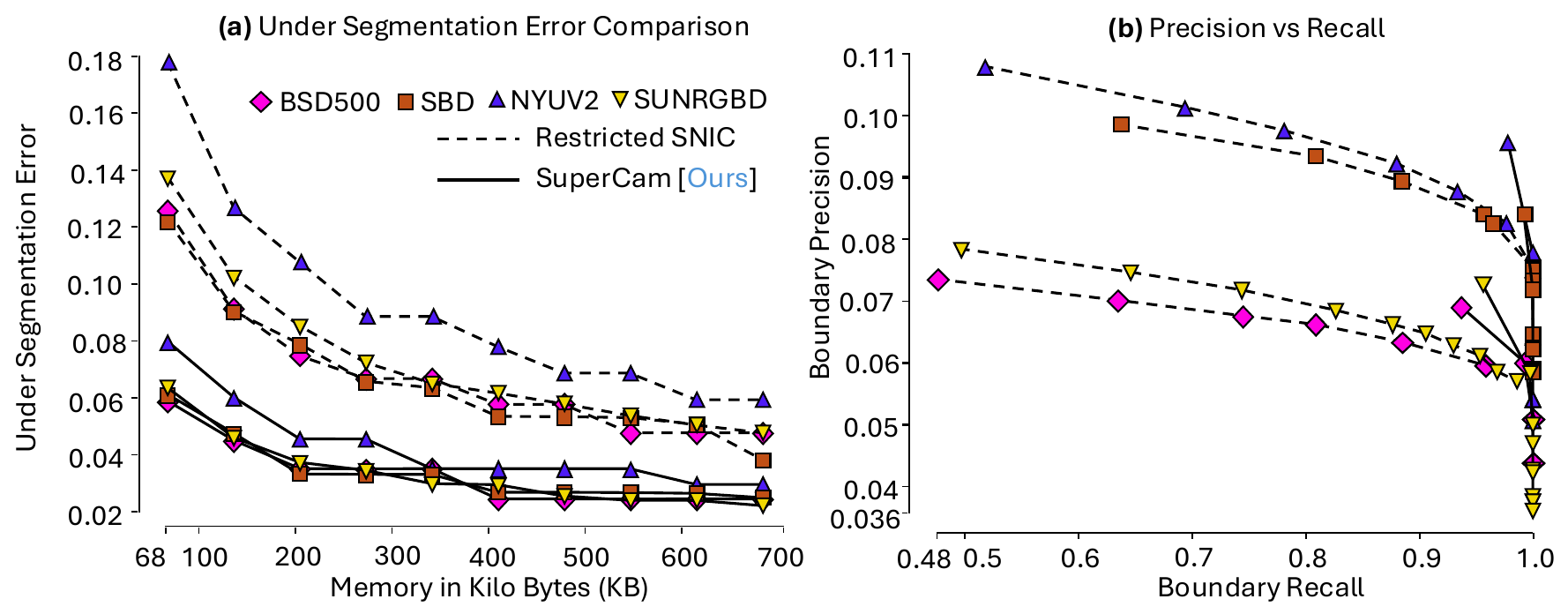}
    \caption{\textbf{Quantitative comparison with non-learning based methods.} We compare the superpixel segmentation quantitatively using the BSD500, NYUV2, SBD and SUNRGBD datasets. \textbf{(a)~Under Segmentation Error} Under Segmentation Error comparison for SuperCam and SNIC for all datasets. SuperCam does at least twice as well as SNIC when using the same amount of memory. \textbf{(b)~Superpixel Performance} Precision vs recall plot for the SuperCam and SNIC algorithms. SuperCam has better recall than SNIC on all datasets, although it has a lower precision than SNIC due to the higher quantity of superpixels used for the same amount of memory.}
    \label{fig:superpixel_eval}
    \vspace{-0.1in}
\end{figure}

\begin{figure*}[!ht]
    \centering
    \includegraphics[width=0.97\linewidth]{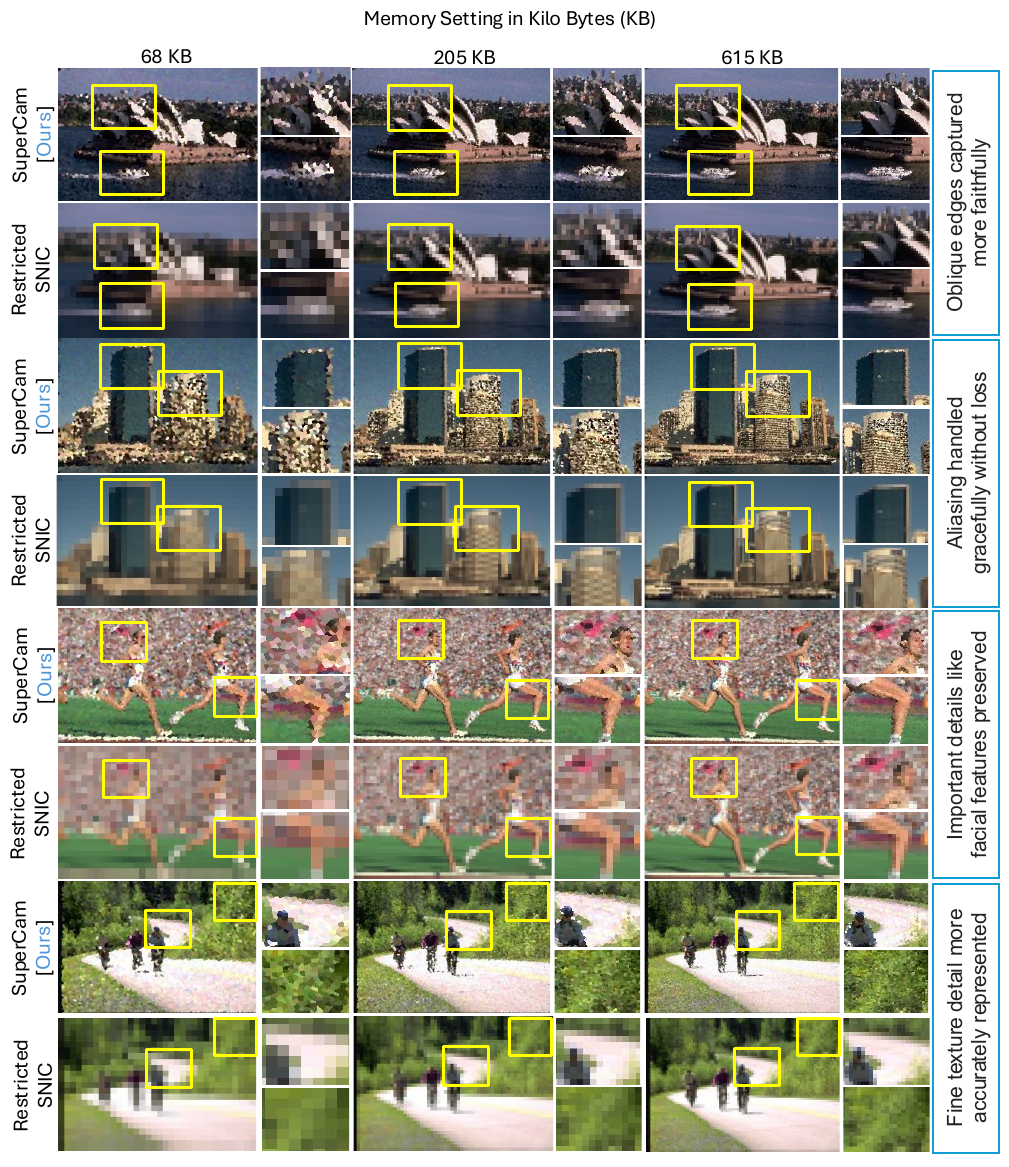}
    \vspace{-0.1in}
    \caption{\textbf{Comparison of memory-constrained SNIC and SuperCam superpixel images.} Superpixel images generated from the BSD500 dataset for different memory settings in kilobytes (KB), shown here \emph{without} Gaussian blur applied for either method. 
    The SuperCam results show higher fidelity, better textural details, and less aliasing.}
    \label{fig:SuperpixelImages}
    \vspace{-0.1in}
\end{figure*}

\begin{figure*}[t!]
    \centering
    \includegraphics[width=\textwidth]{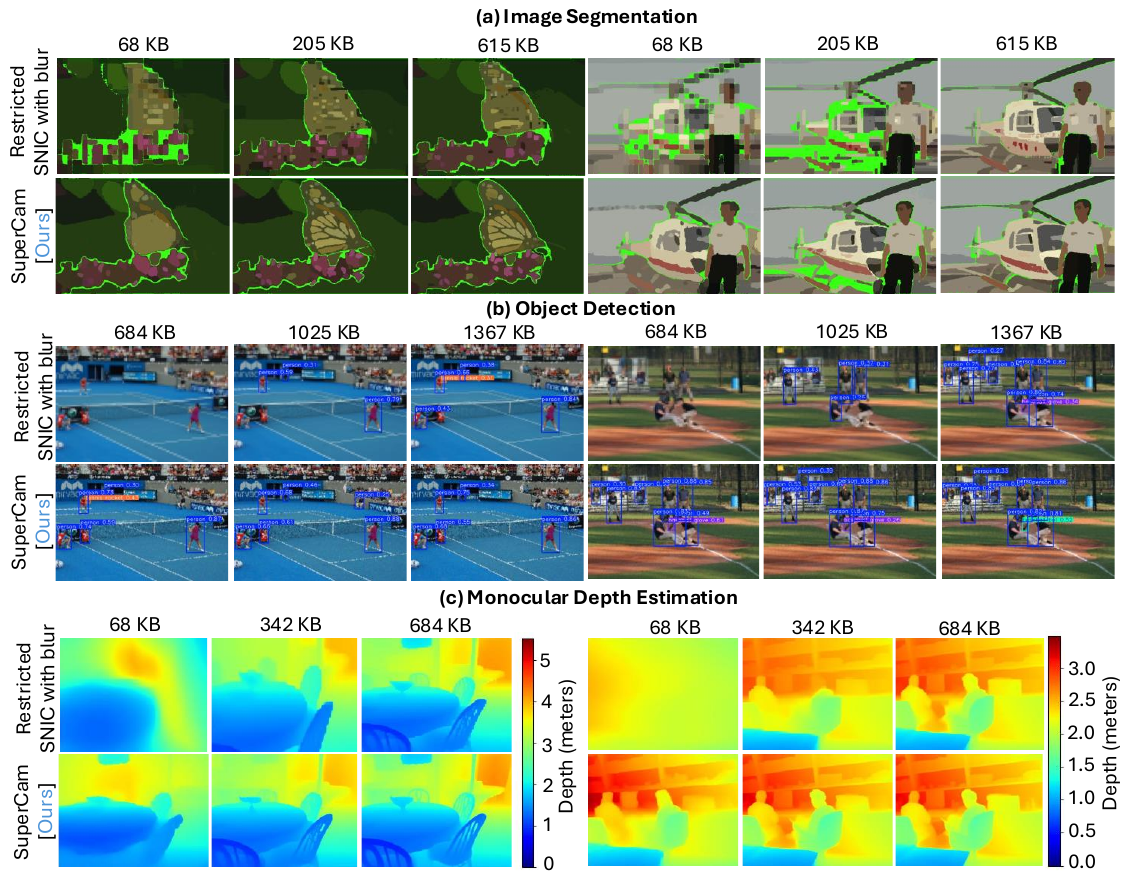}
    \vspace{-0.2in}
    \caption{\textbf{Visual comparison of SuperCam and memory constrained SNIC for three different computer vision tasks.} \textbf{(a)} Image Segmentation results using SAM2 on the BSD500 dataset. The florescent green color refers to regions of the image that were not given any mask by the SAM2 model. \textbf{(b)} Object Detection results using the YOLOv12 model on the COCO dataset. \textbf{(c)} Monocular Depth Estimation results using the DepthAnythingV2 model on the NYUV2 dataset. Rows compare across algorithms and columns show different memory settings (in kilobytes (KB)). SuperCam produces visually better results than SNIC using the same amount of memory.}
    \label{fig:img_seg_results}
    \vspace{-0.2in}
\end{figure*}

\section{Results}
\label{sec:Results}

We show results for four different computer vision tasks spanning low-, mid- and high-level vision.
First, we show quantitative and qualitative evaluation of superpixel segmentation results, evaluating superpixel quality using the standard metrics of under segmentation error and precision vs. recall for those datasets where ground truth (hand-annotated) segmentation is available.
Second, we show the task of image segmentation, where, using the superpixel image as input, a state-of-the-art transformer-based semantic segmentation model is used to generate segmentation masks.
Third, we show results for object detection and localization on superpixel images using standard benchmark datasets.
Fourth, we show results for monocular depth estimation that assigns (relative) depth information at a per-pixel level.
We chose these four tasks because they seem naturally suited for superpixelated input images where fine detail at the level of individual pixels is not available.
Finally, we also show some qualitative results from real-world datasets of binary frames captured using a research-grade SPAD camera \cite{burst_vision}.

\subsection{Superpixel Evaluation}

\noindent\textbf{Comparison with non-learning-based methods.}
We evaluate the quality of SuperCam's superpixel segmentation output using the under segmentation error \cite{UEError} and  boundary precision vs. recall plot.
The precision and recall values are determined using the implementation given in \cite{superpixel-survey1}.
Fig.~\ref{fig:superpixel_eval} shows quantitative evaluation metrics on the BSD500 \cite{BSD500}, NYUV2 \cite{NYUV2}, SBD \cite{SBD} and SUNRGBD \cite{SUNRGBD1, SUNRGBD2, SUNRGBD3, SUNRGBD4} datasets.
Observe that overall trends are similar across the four datasets used.
The under segmentation error decreases monotonically as we increase the number of superpixels.
For the same memory usage, SuperCam performs at least twice as well as SNIC and shows better recall for the same precision values as SNIC. We also provide a comparison with memory restricted SLIC \cite{SLIC} and ERS \cite{ERS} in the supplement, both showing similar performance and error trends as those of SNIC.
Visual comparison results shown in Fig.~\ref{fig:SuperpixelImages} indicate that for applications that are resource-constrained in terms of memory, SuperCam gives better edge preservation and a more complete superpixel image compared to a memory-constrained version of SNIC.

\smallskip
\noindent \textbf{Comparison with learning based methods.}
In the supplement, we compare SuperCam against two recent deep-learning-based superpixel models SPAM \cite{spam} and LNS-Net \cite{LNSNET}. The lowest superpixel/memory setting of SuperCam (68KB) is comparable with LNS-Net using 800 superpixels and $\sim$2 GB memory. We also compare quantized models with smaller memory footprints. SuperCam outperforms on Under Segmentation Error while consuming $>1000 \times$ less memory (kB vs GB).

\begin{figure}[!h]
    \centering
    \includegraphics[width=\linewidth]{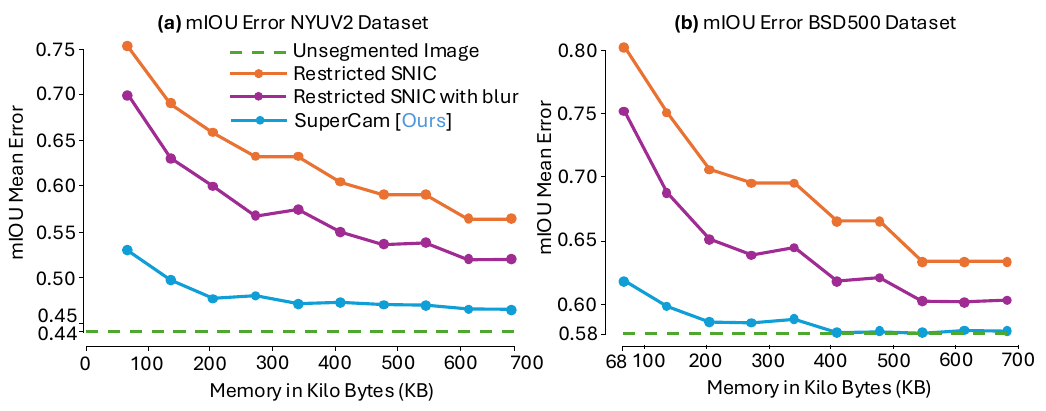}
    \vspace{-0.2in}
    \caption{
    \textbf{Quantitative evaluation of image segmentation.} We compare the quality of image segmentation results produced by the SegmentAnythingV2 model using the mIOU metric on publicly available \textbf{(a)}~NYUV2 and \textbf{(b)}~BSD500 datasets.
    Observe that our proposed SuperCam method achieves a lower mIOU mean error than SNIC.
    The error approaches that of an unsegmented image as the number of superpixels is increased.    
    }
    
    \label{fig:img_seg_eval}
    \vspace{-0.1in}
\end{figure}

\begin{figure}[!ht]
    \centering
    \includegraphics[width=0.90\linewidth]{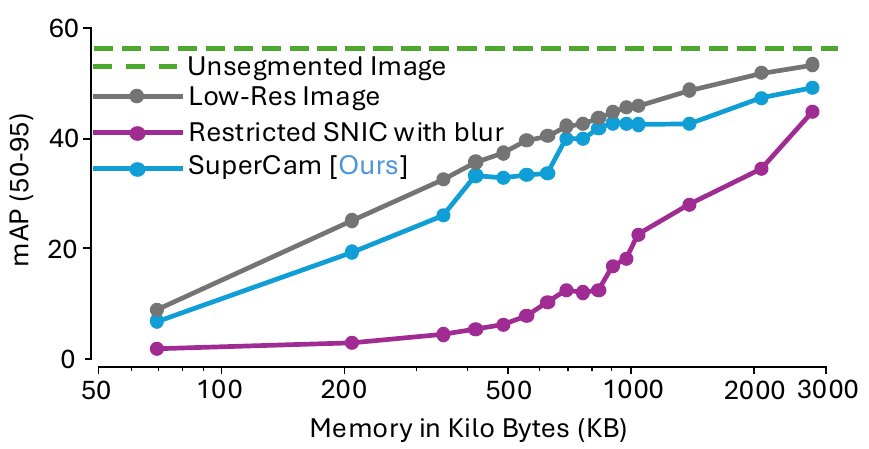}
    \vspace{-0.1in}
    \caption{\textbf{Quantitative evaluation of object detection.} We compare object detection performance for the superpixel images of SNIC and SuperCam with downsampled images. We use mAP(50-95) values for the ground truth bounding boxes produced by the YOLOv12 on the COCO dataset. Observe that SuperCam detections are better than memory constrained SNIC and comparable to results for resolution-matched images. Missed detections in SuperCam are primarily smaller objects lost due to sampling limitations. Both SNIC and SuperCam converge toward the results produced by an unsegmented image as the available memory is increased.}
    \label{fig:obj_det_eval}
    \vspace{-0.2in}
\end{figure}

\begin{figure}[!ht]
    \centering
    \includegraphics[width=\linewidth]{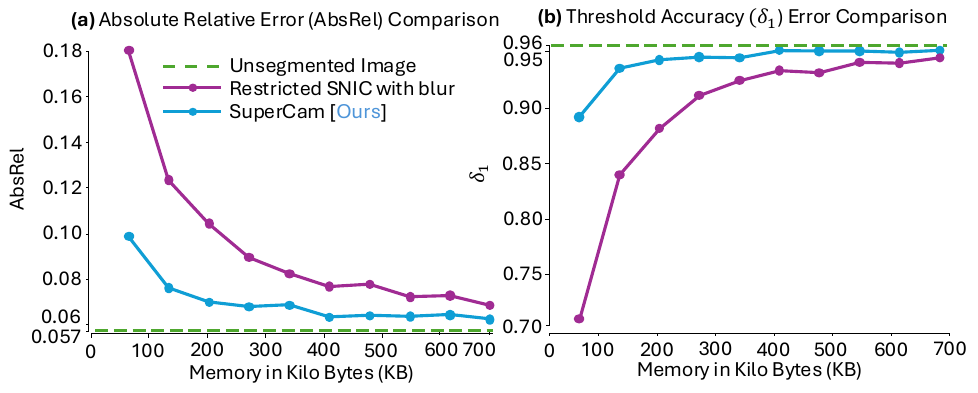}
    \vspace{-0.2in}
    \caption{\textbf{Quantitative evaluation of monocular depth estimation.} We compare the \textbf{(a)}~Absolute Relative Error (AbsRel) and \textbf{(b)}~Threshold Accuracy ($\delta_{1}$) error metrics for depth estimates produced by the DepthAnythingV2 model on the NYUV2 dataset. Observe that SuperCam error metrics are better when compared to SNIC using the same amount of memory. Both algorithms converge to the error values produced by the unsegmented image as memory increases.}
    \label{fig:mono_depth_eval}
    \vspace{-0.2in}
\end{figure}

\begin{figure*}[!ht]
    \centering
    \includegraphics[width=0.95\textwidth]{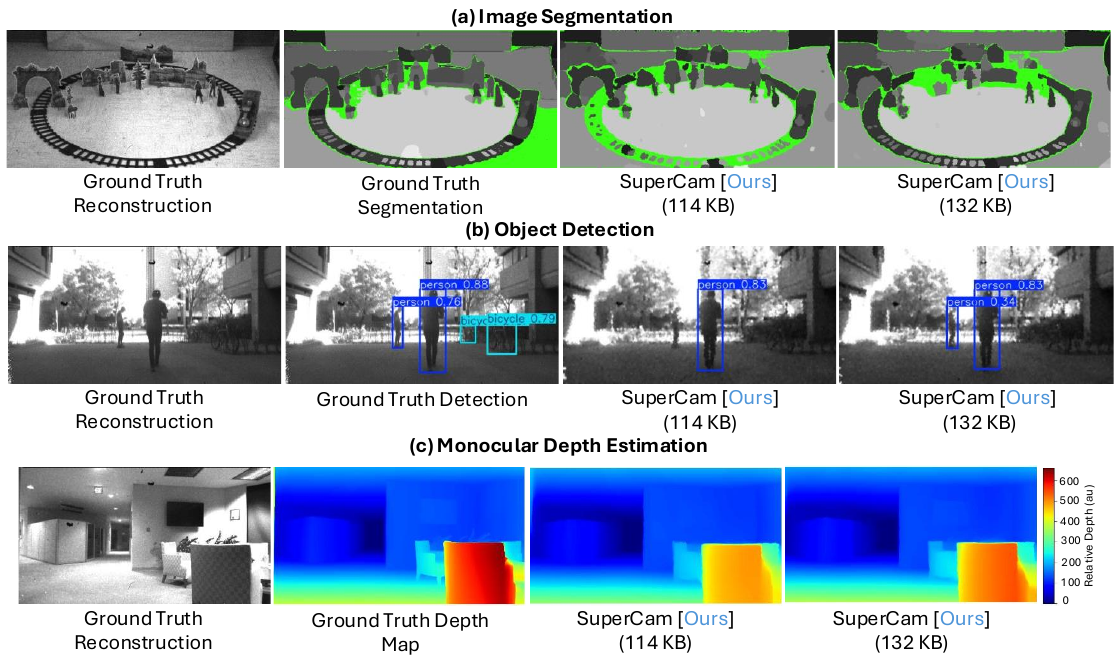}
    \vspace{-0.1in}
    \caption{\textbf{Qualitative comparison of SPAD camera results.} We show experimental results of SuperCam on real world SPAD data from the SWISS SPAD sensor used in the Burst-Vision work. \textbf{(a)} Image Segmentation results using SAM2. The florescent green color refers to regions of the image that were not given any mask by SAM2 model. \textbf{(b)} Object Detection results using the YOLOv12 model. \textbf{(c)} Monocular Depth Estimation results using the DepthAnythingV2 model. Observe that results get better as we increase the memory used.}
    \vspace{-0.1in}
    \label{fig:real_SPAD_results}
    \vspace{-0.1in}
\end{figure*}

\subsection{Image Segmentation}
We use Segment Anything Model 2 \cite{SAM2} to perform image segmentation on the BSD500 \cite{BSD500}, NYUV2 \cite{NYUV2} and SBD \cite{SBD} datasets.
As seen in Fig.~\ref{fig:img_seg_results}(a), SuperCam segmentation results have overall fewer un-segmented regions (shown in fluorescent green) compared to SNIC.
Moreover, our results also show more successful recovery of larger objects, such as the helicopter, even at lower memory levels.
We provide a quantitative evaluation using the mIOU error metric in Fig.~\ref{fig:img_seg_eval} for the NYUV2 \cite{NYUV2} and BSD500 \cite{BSD500} datasets. 
Our method provides a consistently lower mIOU error than SNIC across both datasets.
Moreover, the error metric shows a decreasing trend, which appears to converge to the error achieved if the original unsegmented image were used instead.

\subsection{Object Detection}
We ran an off-the-shelf object detection model, YOLOv12 \cite{yolov12}, on superpixel outputs generated from SuperCam and a memory-constrained version of SNIC for comparison.
We show results for the COCO dataset \cite{COCO}. 
Observe that in Fig.~\ref{fig:img_seg_results}(b)  SuperCam successfully detects and localizes objects even at lower memory settings, while the same objects are completely missed when using SNIC.
Both methods perform comparably at higher memory settings.
A quantitative comparison of the precision of the detected bounding boxes is shown using the $mAP(50-95)$ score in Fig.~\ref{fig:obj_det_eval}.
Our method consistently outperforms SNIC and gracefully approaches the mAP score of an unsegmented image as the number of superpixels (memory) is increased.

We note that superpixel images underperform compared to high-resolution images, for detecting extremely small objects that occupy very few pixels in the original image.
This is a fundamental limitation of the superpixelation approach: applications that require reliable detection of extremely small objects will necessarily have to either trade off extra memory or re-run the superpixelation algorithm on an optically zoomed region of the scene.

\subsection{Monocular Depth Estimation}
Fig.~\ref{fig:img_seg_results}(c) shows visual results for a state-of-the-art monocular depth estimation model, Depth Anything V2 \cite{DepthAnythingv2}, using superpixelated images generated with SNIC and SuperCam.
The predicted depth is compared to the ground truth for the KITTI \cite{KITTI}, DIODE \cite{DIODE}, NYUv2 \cite{NYUV2}, and Sintel \cite{Sintel} datasets. 
Observe that our results consistently outperform SNIC in terms of recovering depth edges for small objects and details.
Some of these details are visible in the SuperCam output even at the lowest memory settings, while SNIC produces an unusable depth map.
As seen in Fig.~\ref{fig:mono_depth_eval} SuperCam consistently outperforms SNIC on quantitative quality metrics such as the absolute relative error (AbsRel) and the threshold accuracy ($\delta_{1}$) error metrics.
Refer to the supplement for more results.

\subsection{Hardware Results}
Finally, we show SuperCam results on a publicly available real-world dataset, \cite{burst_vision} captured using a SwissSPAD sensor \cite{ulku2017512}.
These real-world captures consist of stacks of binary-valued photon detection frames, capturing indoor and outdoor settings over a wide dynamic range of illumination. 
We reconstruct the ground-truth intensity image by summing these binary frames and estimating scene intensity using log-compression as described in Sec.~\ref{sec:ImagingModel}. 
Qualitative results on three of the image sequences are shown in Fig.~\ref{fig:real_SPAD_results}.
The reconstructed images are noisy due to Poisson noise and defective (dead and hot) pixels, and saturated in some regions due to challenging lighting conditions. 
Nevertheless, SuperCam performs comparable to ground truth results for all three computer vision tasks: segmentation, object detection, and monocular depth estimation. 
\section{Conclusion and Future Work}
\label{sec:Conclusion}
This paper introduces a new camera design that produces superpixel images on the fly without storing or capturing the entire image. Our results show that the information encoded in a SuperCam image is sufficient to perform image segmentation, object detection, and depth estimation.
We hope that this work will serve as motivation for further research into the design of cameras that inherently capture parsimonious image representations and bypass the step of forming a high-resolution image altogether.

Future work for SuperCam would be the development of a hardware prototype for the camera. Although synthetic and experimental results look promising, SuperCam still remains a conceptual idea and releasing it in hardware will push the idea further. Additionally, it should be possible to reduce the memory required to implement SuperCam by another factor of two if we were to use a more lightweight data structure that does not include additional information necessary for debugging and testing algorithmic variations. 

{
    \small
    \bibliographystyle{ieeenat_fullname}
    \bibliography{main}

@String(PAMI= {IEEE Transactions on Pattern Analysis Machine Intelligence})

@String(IJCV= {Internation Journal of Computer Vision})

@String(CVPR= {Proceedings of the IEEE/CVF Conference on Computer Vision and Pattern Recognition})

@String(ICCV= {Proceedings of the IEEE/CVF International Conference on Computer Vision})

@String(ECCV= {Proceedings of the European conference on Computer Vision})

@String(NIPS= {Proceedings of the Advances in Neural Information Processing Systems})

@String(BMVC= {Proceedings of the British Machine Vision Conference})

@String(TOG= {ACM Transactions on Graphics})

@String(TIP= {IEEE Transactions on Image Processing})

@String(IROS= {Proceedings of the IEEE/RSJ International Conference on Intelligent Robots and Systems})

@String(ICRA= {Proceedings of the IEEE International Conference on Robotics and Automation})

@String(WACV= {Proceedings of the IEEE/CVF Winter Conference on Applications of Computer Vision})

@String(ICIP= {Proceedings of the IEEE International Conference on Image Processing})

@inproceedings{SNIC,
	location = {Honolulu, {HI}},
	title = {Superpixels and Polygons Using Simple Non-iterative Clustering},
	isbn = {978-1-5386-0457-1},
	url = {http://ieeexplore.ieee.org/document/8100003/},
	doi = {10.1109/CVPR.2017.520},
	eventtitle = {2017 {IEEE} Conference on Computer Vision and Pattern Recognition ({CVPR})},
	pages = {4895--4904},
	booktitle = CVPR,
	author = {Achanta, Radhakrishna and Susstrunk, Sabine},
	urldate = {2025-03-24},
	date = {2017-07},
    year = {2017},
	langid = {english},
}

@inproceedings{superpixel-survey1,
  author    = {David Stutz and Alexander Hermans and Bastian Leibe},
  title     = {Superpixels: An evaluation of the state-of-the-art},
  booktitle   = {Computer Vision and Image Understanding},
  volume    = {166},
  pages     = {1--27},
  year      = {2018},
  url       = {https://doi.org/10.1016/j.cviu.2017.03.007},
  doi       = {10.1016/j.cviu.2017.03.007},
}

@inproceedings{superpixel-survey2,
title={A comprehensive review and new taxonomy on superpixel segmentation},
  author={Barcelos, Isabela Borlido and Belem, Felipe De Castro and Joao, Leonardo De Melo and Patroc{\'\i}nio Jr, Zenilton KG Do and Falcao, Alexandre Xavier and Guimar{\~a}es, Silvio Jamil Ferzoli},
  booktitle={ACM Computing Surveys},
  volume={56},
  number={8},
  pages={1--39},
  year={2024}
}

@inproceedings{sem_seg_application1,
  title={Improving semantic image segmentation with a probabilistic superpixel-based dense conditional random field},
  author={Zhang, Liang and Li, Huan and Shen, Peiyi and Zhu, Guangming and Song, Juan and Shah, Syed Afaq Ali and Bennamoun, Mohammed and Zhang, Li},
  booktitle={IEEE Access},
  volume={6},
  pages={15297--15310},
  year={2018},
}

@inproceedings{sem_seg_application2,
  title={Adaptive superpixel for active learning in semantic segmentation},
  author={Kim, Hoyoung and Oh, Minhyeon and Hwang, Sehyun and Kwak, Suha and Ok, Jungseul},
  booktitle=ICCV,
  pages={943--953},
  year={2023}
}

@inproceedings{sem_seg_application3,
  title={Semantic segmentation from sparse labeling using multi-level superpixels},
  author={Alonso, I{\~n}igo and Murillo, Ana C},
  booktitle=IROS,
  pages={5785--5792},
  year={2018}
}

@inproceedings{sem_seg_application4,
  title={Viewal: Active learning with viewpoint entropy for semantic segmentation},
  author={Siddiqui, Yawar and Valentin, Julien and Nie{\ss}ner, Matthias},
  booktitle=CVPR,
  pages={9433--9443},
  year={2020}
}

@inproceedings{sem_seg_application5,
  title={Superpixel transformers for efficient semantic segmentation},
  author={Zhu, Alex Zihao and Mei, Jieru and Qiao, Siyuan and Yan, Hang and Zhu, Yukun and Chen, Liang-Chieh and Kretzschmar, Henrik},
  booktitle=IROS,
  pages={7651--7658},
  year={2023}
}

@article{3d_recon_application1,
  title={Multi-view superpixel stereo in urban environments},
  author={Mi{\v{c}}u{\v{s}}{\'\i}k, Branislav and Ko{\v{s}}eck{\'a}, Jana},
  journal=IJCV,
  volume={89},
  number={1},
  pages={106--119},
  year={2010}
}

@inproceedings{3d_recon_application2,
  title={Real-time local 3D reconstruction for aerial inspection using superpixel expansion},
  author={Teixeira, Lucas and Chli, Margarita},
  booktitle=ICRA,
  pages={4560--4567},
  year={2017}
}

@article{3d_recon_application3,
  title={Superpixel soup: Monocular dense 3d reconstruction of a complex dynamic scene},
  author={Kumar, Suryansh and Dai, Yuchao and Li, Hongdong},
  journal=PAMI,
  volume={43},
  number={5},
  pages={1705--1717},
  year={2019}
}

@inproceedings{3d_recon_application4,
  title={Superpixel meshes for fast edge-preserving surface reconstruction},
  author={B{\'o}dis-Szomor{\'u}, Andr{\'a}s and Riemenschneider, Hayko and Van Gool, Luc},
  booktitle=CVPR,
  pages={2011--2020},
  year={2015}
}

@inproceedings{obj_det_application1,
  title={Improving an object detector and extracting regions using superpixels},
  author={Shu, Guang and Dehghan, Afshin and Shah, Mubarak},
  booktitle=CVPR,
  pages={3721--3727},
  year={2013}
}

@inproceedings{obj_det_application2,
  title={Object detection by labeling superpixels},
  author={Yan, Junjie and Yu, Yinan and Zhu, Xiangyu and Lei, Zhen and Li, Stan Z},
  booktitle=CVPR,
  pages={5107--5116},
  year={2015}
}

@article{obj_det_application3,
  title={Automatic pavement object detection using superpixel segmentation combined with conditional random field},
  author={Sultani, Waqas and Mokhtari, Soroush and Yun, Hae-Bum},
  journal={IEEE Transactions on Intelligent Transportation Systems},
  volume={19},
  number={7},
  pages={2076--2085},
  year={2017}
}

@article{obj_det_application4,
  title={Object detection using SURF and superpixels},
  author={Lopez-de-la-Calleja, Miriam and Nagai, Takayuki and Attamimi, Muhammad and Nakano-Miyatake, Mariko and Perez-Meana, Hector and others},
  journal={Journal of Software Engineering and Applications},
  volume={6},
  number={09},
  pages={511},
  year={2013}
}

@inproceedings{depth_application1,
  title={Improving Depth Estimation Using Superpixels.},
  author={Cambra, Ana B and Mu{\~n}oz, Adolfo and Murillo, Ana C and Guerrero, Jos{\'e} Jes{\'u}s and Gutierrez, Diego},
  booktitle={CEIG},
  pages={49--58},
  year={2014}
}

@article{depth_application2,
  title={Fast and accurate depth estimation from sparse light fields},
  author={Chuchvara, Aleksandra and Barsi, Attila and Gotchev, Atanas},
  journal=TIP,
  volume={29},
  pages={2492--2506},
  year={2019}
}

@inproceedings{depth_application3,
  title={A dense depth estimation method using superpixels},
  author={Jin, Feng and Li, Xuefeng},
  booktitle={Proceedings of the International Computer Conference on Wavelet Active Media Technology and Information Processing},
  pages={290--294},
  year={2015}
}

@article{depth_application4,
  title={Adaptive illumination based depth sensing using deep superpixel and soft sampling approximation},
  author={Dai, Qiqin and Li, Fengqiang and Cossairt, Oliver and Katsaggelos, Aggelos K},
  journal={IEEE Transactions on Computational Imaging},
  volume={8},
  pages={224--235},
  year={2022}
}

@inproceedings{obj_track_application1,
  title={Superpixel tracking},
  author={Wang, Shu and Lu, Huchuan and Yang, Fan and Yang, Ming-Hsuan},
  booktitle=ICCV,
  pages={1323--1330},
  year={2011}
}

@article{obj_track_application2,
  title={Robust superpixel tracking},
  author={Yang, Fan and Lu, Huchuan and Yang, Ming-Hsuan},
  journal=TIP,
  volume={23},
  number={4},
  pages={1639--1651},
  year={2014}
}

@inproceedings{obj_track_application3,
  title={Tracking using superpixel features},
  author={Jingjing, Liu and Ying, Chen and Cheng, Zha and Hua, Yu and Li, Zhao},
  booktitle={Proceedings of the International Conference on Measuring Technology and Mechatronics Automation},
  pages={878--881},
  year={2016}
}

@article{obj_track_application4,
  title={Constrained superpixel tracking},
  author={Wang, Lijun and Lu, Huchuan and Yang, Ming-Hsuan},
  journal={IEEE Transactions on Cybernetics},
  volume={48},
  number={3},
  pages={1030--1041},
  year={2017}
}

@article{opt_flow_application1,
  title={Highly accurate optical flow estimation on superpixel tree},
  author={Hu, Yinlin and Song, Rui and Li, Yunsong and Rao, Peng and Wang, Yangli},
  journal={Image and Vision Computing},
  volume={52},
  pages={167--177},
  year={2016},
  publisher={Elsevier}
}

@inproceedings{opt_flow_application2,
  title={Moving object detection in dynamic scenes based on optical flow and superpixels},
  author={Li, Xiuzhi and Xu, Chuanluo},
  booktitle={Proceedings of the IEEE International Conference on Robotics and Biomimetics},
  pages={84--89},
  year={2015}
}

@article{opt_flow_application3,
  title={Full view optical flow estimation leveraged from light field superpixel},
  author={Zhu, Hao and Sun, Xiaoming and Zhang, Qi and Wang, Qing and Robles-Kelly, Antonio and Li, Hongdong and You, Shaodi},
  journal={IEEE Transactions on Computational Imaging},
  volume={6},
  pages={12--23},
  year={2019}
}

@inproceedings{opt_flow_application4,
  title={Guiding optical flow estimation using superpixels},
  author={Gkamas, Theodosios and Nikou, Christophoros},
  booktitle={Proceedings of the International Conference on Digital Signal Processing},
  pages={1--6},
  year={2011}
}

@inproceedings{AINET,
  title={AINet: Association implantation for superpixel segmentation},
  author={Wang, Yaxiong and Wei, Yunchao and Qian, Xueming and Zhu, Li and Yang, Yi},
  booktitle=ICCV,
  pages={7078--7087},
  year={2021}
}

@inproceedings{DAL-HERS,
  title={Hers superpixels: Deep affinity learning for hierarchical entropy rate segmentation},
  author={Peng, Hankui and Aviles-Rivero, Angelica I and Sch{\"o}nlieb, Carola-Bibiane},
  booktitle=WACV,
  pages={217--226},
  year={2022}
}

@inproceedings{LNSNET,
  title={Learning the superpixel in a non-iterative and lifelong manner},
  author={Zhu, Lei and She, Qi and Zhang, Bin and Lu, Yanye and Lu, Zhilin and Li, Duo and Hu, Jie},
  booktitle=CVPR,
  pages={1225--1234},
  year={2021}
}

@misc{wiredarticle,
  title = {{Square Pixel Inventor Tries to Smooth Things Out}},
  author={Ehrenberg, Rachel},
  year={2018},
  publisher={Wired},
  note="\url{https://www.wired.com/2010/06/smoothing-square-pixels/} (last accessed Nov 10, 2025)"
}

@article{kuroda1982position,
  title={Position sensitive photomultiplier},
  author={Kuroda, K},
  journal={Nuclear Instruments and Methods in Physics Research},
  volume={196},
  number={1},
  pages={187--197},
  year={1982},
  publisher={Elsevier}
}

@article{gibson2020single,
  title={Single-pixel imaging 12 years on: a review},
  author={Gibson, Graham M and Johnson, Steven D and Padgett, Miles J},
  journal={Optics express},
  volume={28},
  number={19},
  pages={28190--28208},
  year={2020},
  publisher={OSA}
}

@inproceedings{dierickx1996random,
  title={Random addressable active pixel image sensors},
  author={Dierickx, Bart and Scheffer, Danny and Meynants, Guy and Ogiers, Werner and Vlummens, Jan},
  booktitle={Advanced Focal Plane Arrays and Electronic Cameras},
  volume={2950},
  pages={2--7},
  year={1996},
  organization={SPIE}
}

@article{SLIC,
  title={SLIC superpixels compared to state-of-the-art superpixel methods},
  author={Achanta, Radhakrishna and Shaji, Appu and Smith, Kevin and Lucchi, Aurelien and Fua, Pascal and S{\"u}sstrunk, Sabine},
  journal=PAMI,
  volume={34},
  number={11},
  pages={2274--2282},
  year={2012}
}

@inproceedings{SEEDS,
  title={Seeds: Superpixels extracted via energy-driven sampling},
  author={Van den Bergh, Michael and Boix, Xavier and Roig, Gemma and De Capitani, Benjamin and Van Gool, Luc},
  booktitle=ECCV,
  pages={13--26},
  year={2012}
}

@inproceedings{ERS,
  title={Entropy rate superpixel segmentation},
  author={Liu, Ming-Yu and Tuzel, Oncel and Ramalingam, Srikumar and Chellappa, Rama},
  booktitle=CVPR,
  pages={2097--2104},
  year={2011}
}

@inproceedings{CRS,
  title={Contour-relaxed superpixels},
  author={Conrad, Christian and Mertz, Matthias and Mester, Rudolf},
  booktitle={Proceedings of the International Workshop on Energy Minimization Methods in Computer Vision and Pattern Recognition},
  pages={280--293},
  year={2013}
}

@article{quanta_burst,
  title={Quanta burst photography},
  author={Ma, Sizhuo and Gupta, Shantanu and Ulku, Arin C and Bruschini, Claudio and Charbon, Edoardo and Gupta, Mohit},
  journal=TOG,
  volume={39},
  number={4},
  pages={79--1},
  year={2020}
}

@article{poission_stat,
  title={Bits from photons: Oversampled image acquisition using binary poisson statistics},
  author={Yang, Feng and Lu, Yue M and Sbaiz, Luciano and Vetterli, Martin},
  journal=TIP,
  volume={21},
  number={4},
  pages={1421--1436},
  year={2011}
}

@inproceedings{UEError,
  title={Superpixel benchmark and comparison},
  author={Neubert, Peer and Protzel, Peter},
  booktitle={Proceedings Forum Bildverarbeitung},
  volume={6},
  pages={1--12},
  year={2012}
}

@inproceedings{BSD500,
  title={A database of human segmented natural images and its application to evaluating segmentation algorithms and measuring ecological statistics},
  author={Martin, David and Fowlkes, Charless and Tal, Doron and Malik, Jitendra},
  booktitle=ICCV,
  volume={2},
  pages={416--423},
  year={2001}
}

@inproceedings{NYUV2,
  title={Indoor segmentation and support inference from rgbd images},
  author={Silberman, Nathan and Hoiem, Derek and Kohli, Pushmeet and Fergus, Rob},
  booktitle=ECCV,
  pages={746--760},
  year={2012}
}

@inproceedings{SBD,
  title={Decomposing a scene into geometric and semantically consistent regions},
  author={Gould, Stephen and Fulton, Richard and Koller, Daphne},
  booktitle=ICCV,
  pages={1--8},
  year={2009}
}

@inproceedings{SUNRGBD1,
 author = {Zhou, Bolei and Lapedriza, Agata and Xiao, Jianxiong and Torralba, Antonio and Oliva, Aude},
 booktitle = NIPS,
 editor = {Z. Ghahramani and M. Welling and C. Cortes and N. Lawrence and K.Q. Weinberger},
 pages = {},
 title = {Learning Deep Features for Scene Recognition using Places Database},
 url = {https://proceedings.neurips.cc/paper_files/paper/2014/file/19ea3982b415d7bb3363917eb3d60c4a-Paper.pdf},
 volume = {27},
 year = {2014}
}

@InProceedings{SUNRGBD2,
author="Silberman, Nathan
and Hoiem, Derek
and Kohli, Pushmeet
and Fergus, Rob",
editor="Fitzgibbon, Andrew
and Lazebnik, Svetlana
and Perona, Pietro
and Sato, Yoichi
and Schmid, Cordelia",
title="Indoor Segmentation and Support Inference from RGBD Images",
booktitle=ECCV,
year="2012",
address="Berlin, Heidelberg",
pages="746--760",
isbn="978-3-642-33715-4"
}

@INPROCEEDINGS{SUNRGBD3,
  author={Janoch, Allison and Karayev, Sergey and Yangqing Jia and Barron, Jonathan T. and Fritz, Mario and Saenko, Kate and Darrell, Trevor},
  booktitle={Proceedings of the IEEE International Conference on Computer Vision Workshops}, 
  title={A category-level 3-D object dataset: Putting the Kinect to work}, 
  year={2011},
  volume={},
  number={},
  pages={1168-1174},
  doi={10.1109/ICCVW.2011.6130382},
  ISSN={},
  month={Nov},}

@INPROCEEDINGS{SUNRGBD4,
  author={Xiao, Jianxiong and Owens, Andrew and Torralba, Antonio},
  booktitle=ICCV, 
  title={SUN3D: A Database of Big Spaces Reconstructed Using SfM and Object Labels}, 
  year={2013},
  volume={},
  number={},
  pages={1625-1632},
  doi={10.1109/ICCV.2013.458},
  ISSN={2380-7504},
  month={Dec},}

@article{SAM2,
  title={SAM 2: Segment Anything in Images and Videos},
  author={Ravi, Nikhila and Gabeur, Valentin and Hu, Yuan-Ting and Hu, Ronghang and Ryali, Chaitanya and Ma, Tengyu and Khedr, Haitham and R{\"a}dle, Roman and Rolland, Chloe and Gustafson, Laura and Mintun, Eric and Pan, Junting and Alwala, Kalyan Vasudev and Carion, Nicolas and Wu, Chao-Yuan and Girshick, Ross and Doll{\'a}r, Piotr and Feichtenhofer, Christoph},
  journal={arXiv preprint arXiv:2408.00714},
  url={https://arxiv.org/abs/2408.00714},
  year={2024}
}

@article{yolov12,
  title={YOLOv12: Attention-Centric Real-Time Object Detectors},
  author={Tian, Yunjie and Ye, Qixiang and Doermann, David},
  journal={arXiv preprint arXiv:2502.12524},
  year={2025}
}

@article{DepthAnythingv2,
  title={Depth anything v2},
  author={Yang, Lihe and Kang, Bingyi and Huang, Zilong and Zhao, Zhen and Xu, Xiaogang and Feng, Jiashi and Zhao, Hengshuang},
  journal=NIPS,
  volume={37},
  pages={21875--21911},
  year={2024}
}

@inproceedings{COCO,
  title={Microsoft coco: Common objects in context},
  author={Lin, Tsung-Yi and Maire, Michael and Belongie, Serge and Hays, James and Perona, Pietro and Ramanan, Deva and Doll{\'a}r, Piotr and Zitnick, C Lawrence},
  booktitle=ECCV,
  pages={740--755},
  year={2014}
}

@article{KITTI,
  title={Vision meets robotics: The kitti dataset},
  author={Geiger, Andreas and Lenz, Philip and Stiller, Christoph and Urtasun, Raquel},
  journal={The International Journal of Robotics Research},
  volume={32},
  number={11},
  pages={1231--1237},
  year={2013}
}

@article{DIODE,
  title={{DIODE}: {A} {D}ense {I}ndoor and {O}utdoor {DE}pth {D}ataset},
  author={Igor Vasiljevic and Nick Kolkin and Shanyi Zhang and Ruotian Luo and
  Haochen Wang and Falcon Z. Dai and Andrea F. Daniele and Mohammadreza Mostajabi and
  Steven Basart and Matthew R. Walter and Gregory Shakhnarovich},
  journal={CoRR},
  volume={abs/1908.00463},
  year={2019},
  url={http://arxiv.org/abs/1908.00463}
}

@inproceedings{Sintel,
  title={A naturalistic open source movie for optical flow evaluation},
  author={Butler, Daniel J and Wulff, Jonas and Stanley, Garrett B and Black, Michael J},
  booktitle=ECCV,
  pages={611--625},
  year={2012}
}

@InProceedings{graph_based_sp1,
  title={Superpixels and supervoxels in an energy optimization framework},
  author={Veksler, Olga and Boykov, Yuri and Mehrani, Paria},
  booktitle=ECCV,
  pages={211--224},
  year={2010}
}

@ARTICLE{TurboPixels,
  title={Turbopixels: Fast superpixels using geometric flows},
  author={Levinshtein, Alex and Stere, Adrian and Kutulakos, Kiriakos N and Fleet, David J and Dickinson, Sven J and Siddiqi, Kaleem},
  journal=PAMI,
  volume={31},
  number={12},
  pages={2290--2297},
  year={2009}
}

@inproceedings{semasuperpixel,
  title={Semasuperpixel: A multi-channel probability-driven superpixel segmentation method},
  author={Wang, Xuehui and Zhao, Qingyun and Fan, Lei and Zhao, Yuzhi and Wang, Tiantian and Yan, Qiong and Chen, Long},
  booktitle=ICIP,
  pages={1859--1863},
  year={2021}
}

@inproceedings{SSN,
  title={Superpixel sampling networks},
  author={Jampani, Varun and Sun, Deqing and Liu, Ming-Yu and Yang, Ming-Hsuan and Kautz, Jan},
  booktitle=ECCV,
  pages={352--368},
  year={2018}
}

@article{E2E-SIS,
  title={End-to-end trainable network for superpixel and image segmentation},
  author={Wang, Kai and Li, Liang and Zhang, Jiawan},
  journal={Pattern Recognition Letters},
  volume={140},
  pages={135--142},
  year={2020},
  publisher={Elsevier}
}

@article{SEEK,
  title={SEEK: A Framework of Superpixel Learning with CNN Features for Unsupervised Segmentation},
  author={Ilyas, Talha and Khan, Abbas and Umraiz, Muhammad and Kim, Hyongsuk},
  journal={Electronics},
  volume={9},
  number={3},
  pages={383},
  year={2020},
  publisher={Multidisciplinary Digital Publishing Institute}
}

@inproceedings{burst_vision,
  title={Burst vision using single-photon cameras},
  author={Ma, Sizhuo and Mos, Paul and Charbon, Edoardo and Gupta, Mohit},
  booktitle=WACV,
  pages={5375--5385},
  year={2023}
}

@inproceedings{sundar2023sodacam,
  title={Sodacam: Software-defined cameras via single-photon imaging},
  author={Sundar, Varun and Ardelean, Andrei and Swedish, Tristan and Bruschini, Claudio and Charbon, Edoardo and Gupta, Mohit},
  booktitle=ICCV,
  pages={8165--8176},
  year={2023}
}

@inproceedings{freeform_pixels,
  title={Minimalist vision with freeform pixels},
  author={Klotz, Jeremy and Nayar, Shree K},
  booktitle=ECCV,
  pages={329--346},
  year={2024}
}

@inproceedings{bayesian_superpixels,
  title={Bayesian-Inspired Space-Time Superpixels},
  author={Gauen, Kent and Chan, Stanley},
  booktitle=ICCV,
  pages={5382--5391},
  year={2025}
}

@inproceedings{CLUSTSEG,
author = {Liang, James and Zhou, Tianfei and Liu, Dongfang and Wang, Wenguan},
title = {CLUSTSEG: clustering for universal segmentation},
year = {2023},
booktitle = {Proceedings of the International Conference on Machine Learning},
articleno = {857},
numpages = {23},
location = {Honolulu, Hawaii, USA},
series = {ICML'23}
}

@inproceedings{zhang2025focal,
  title={{Focal Plane Visual Feature Generation and Matching on a Pixel Processor Array}},
  author={Zhang, Hongyi and Bose, Laurie and Chen, Jianing and Dudek, Piotr and Mayol-Cuevas, Walterio},
  booktitle=ICCV,
  pages={29031--29039},
  year={2025}
}

@inproceedings{bose2025descriptor,
  title={Descriptor-In-Pixel: Point-Feature Tracking For Pixel Processor Arrays},
  author={Bose, Laurie and Chen, Jianing and Dudek, Piotr},
  booktitle=CVPR,
  pages={5392--5400},
  year={2025}
}

@inproceedings{BASS,
  title={Bayesian adaptive superpixel segmentation},
  author={Uziel, Roy and Ronen, Meitar and Freifeld, Oren},
  booktitle=ICCV,
  pages={8470--8479},
  year={2019}
 }

@article{ulku2017512,
  title={{A 512$\times$512 SPAD image sensor with built-in gating for phasor based real-time siFLIM}},
  author={Ulku, Arin Can and Bruschini, Claudio and Michalet, Xavier and Weiss, Shimon and Charbon, EA},
  journal={Proceedings of the International Image Sensors Workshop},
  pages={234--237},
  year={2017}
}

@article{duarte2008single,
  title={Single-pixel imaging via compressive sampling},
  author={Duarte, Marco F and Davenport, Mark A and Takhar, Dharmpal and Laska, Jason N and Sun, Ting and Kelly, Kevin F and Baraniuk, Richard G},
  journal={IEEE Signal Processing Magazine},
  volume={25},
  number={2},
  pages={83--91},
  year={2008},
  publisher={IEEE}
}

@article{ardelean2023computational,
  title={Computational imaging {SPAD} cameras},
  author={Ardelean, Andrei},
  journal={PhD thesis},
  year={2023},
  publisher={{\'E}cole polytechnique f{\'e}d{\'e}rale de Lausanne}
}

@article{gallego2020event,
  title={Event-based vision: A survey},
  author={Gallego, Guillermo and Delbr{\"u}ck, Tobi and Orchard, Garrick and Bartolozzi, Chiara and Taba, Brian and Censi, Andrea and Leutenegger, Stefan and Davison, Andrew J and Conradt, J{\"o}rg and Daniilidis, Kostas and others},
  journal=PAMI,
  volume={44},
  number={1},
  pages={154--180},
  year={2020}
}

@inproceedings{spam,
  title={Superpixel Anything: A general object-based framework for accurate yet regular superpixel segmentation},
  author={Walther, Julien and Giraud, Rémi and Clément, Michaël},
  booktitle = BMVC,
  pages={1035},
  year={2025}
}
}

\clearpage
\onecolumn

\renewcommand{\figurename}{Supplementary Figure}
\renewcommand{\thesection}{Supplementary Note \arabic{section}}
\renewcommand{\theequation}{S\arabic{equation}}
\setcounter{figure}{0}
\setcounter{section}{0}
\setcounter{subsection}{0}
\setcounter{equation}{0}
\setcounter{page}{1}

\begin{center}
\huge Supplementary Document for\\[0.2cm]
\huge ``\mytitle'' \\[1.1cm]
\normalsize Sasidharan Mahalingam, Rachel Brown, Atul Ingle\\[0.2cm]
\end{center}
\setcounter{page}{1}



\begin{figure*}[hbt!]
    \centering
    \includegraphics[width=0.9\linewidth]{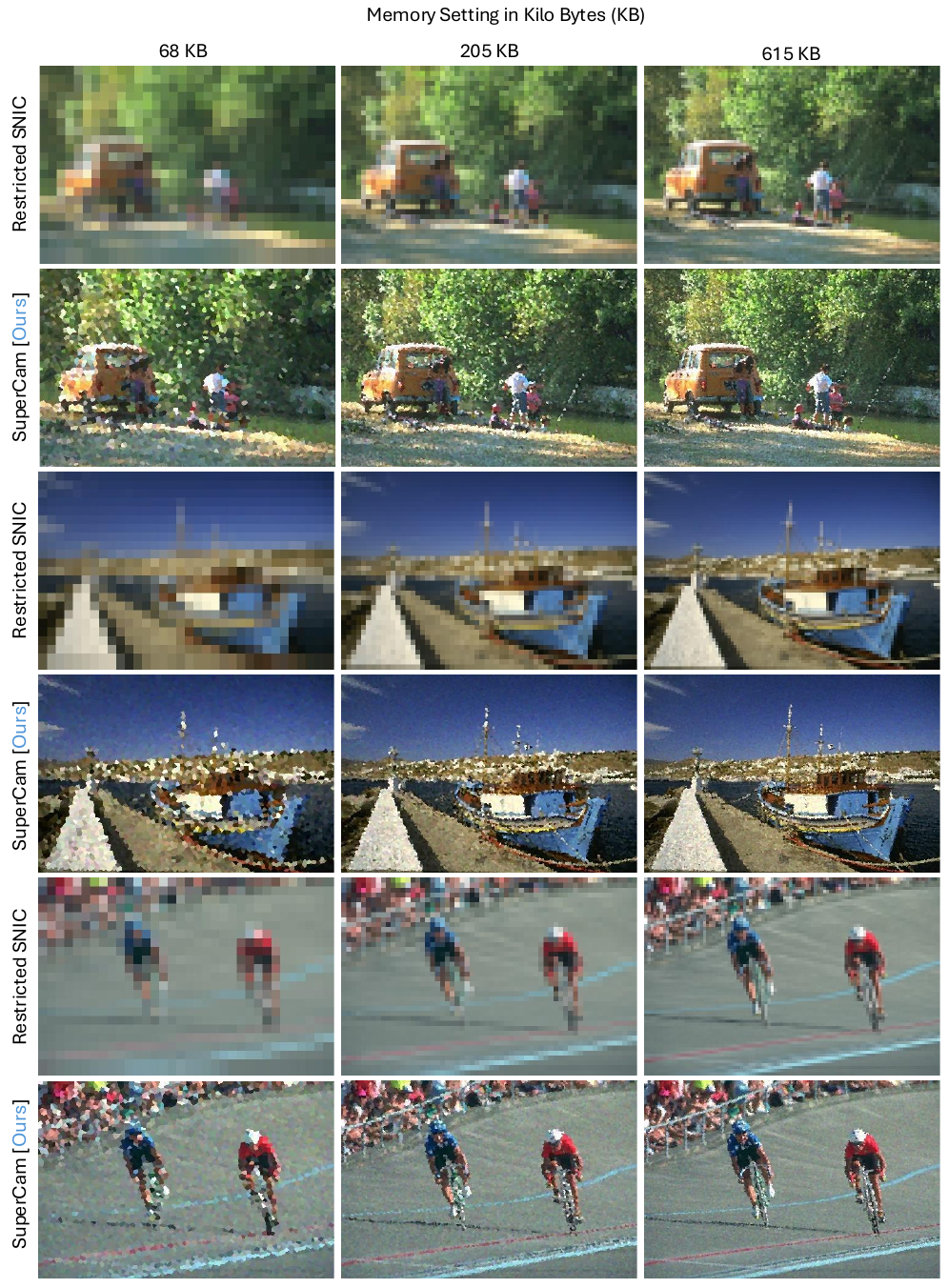}
    \caption{\textbf{Comparison of memory-restricted SNIC and SuperCam superpixel images.} Images taken from the BSD500 dataset, shown at different memory settings in kilobytes (KB).  These are raw output images \emph{without} Gaussian blur applied. 
    SuperCam results show better visual details and less aliasing.}
    \label{fig:Superpixel_Images_Supplement}
\end{figure*}

\section{Superpixel Segmentation Results}
\label{sec:Results_Comparison_Supplementary}

Suppl. Fig. \ref{fig:Superpixel_Images_Supplement} shows more qualitative visual comparisons generated using SuperCam and memory restricted SNIC~\cite{SNIC} at varying memory budgets, shown here \emph{without} Gaussian blur applied.

In the remaining supplement sections we provide additional qualitative and quantitative evaluations, including comparisons with additional superpixel algorithms and results for three computer vision tasks: image segmentation, object detection and monocular depth estimation. We also provide mathematical definitions for the quantitative error metrics that we used and additional detail regarding how we derived the optimal Gaussian blur kernel.

\clearpage \newpage

\section{Comparisons with other superpixel algorithms}
\label{sec:Comparisons_Supplement}
\begin{figure}[!h]
    \centering
    \includegraphics[width=0.5\linewidth]{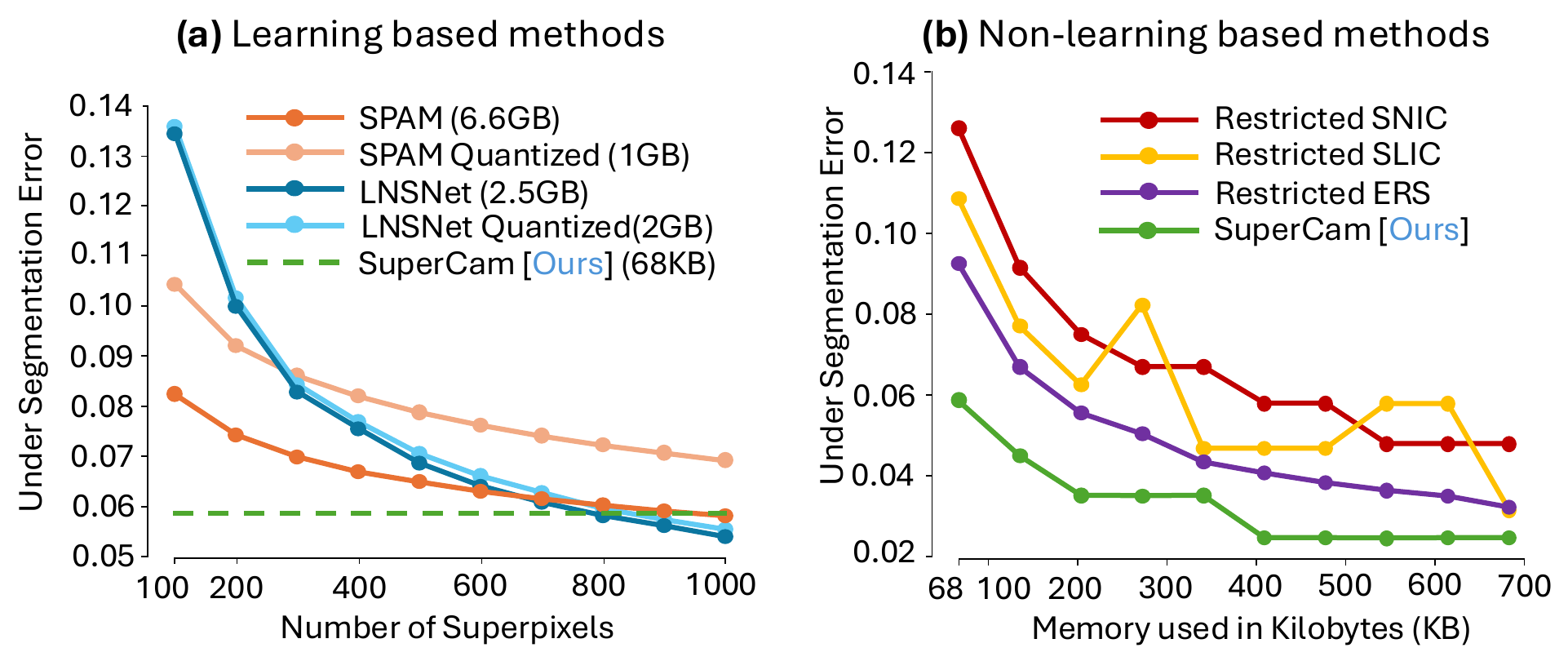}
    \caption{\textbf{Comparisons of SuperCam with other superpixel algorithms.} Quantitative comparisons of SuperCam with other learning based and non-learning based superpixel algorithms. \textbf{(a) Comparisons of SuperCam with learning based superpixel algorithms:} We show comparisons of SuperCam with two recent learning based algorithms SPAM and LNS-Net. \textbf{(b) Comparisons of SuperCam with non-learning based superpixel algorithms:} Here we show comparisons of SuperCam with memory restricted SNIC, SLIC and ERS on the BSD500 dataset. We can see that the overall trend is similar for all the memory restricted algorithms and SuperCam performs better that all algorithms.}
    \label{fig:Superpixel_UE_Comparisons_Supplement}
\end{figure}

\begin{figure}[!h]
    \centering
    \includegraphics[width=0.78\linewidth]{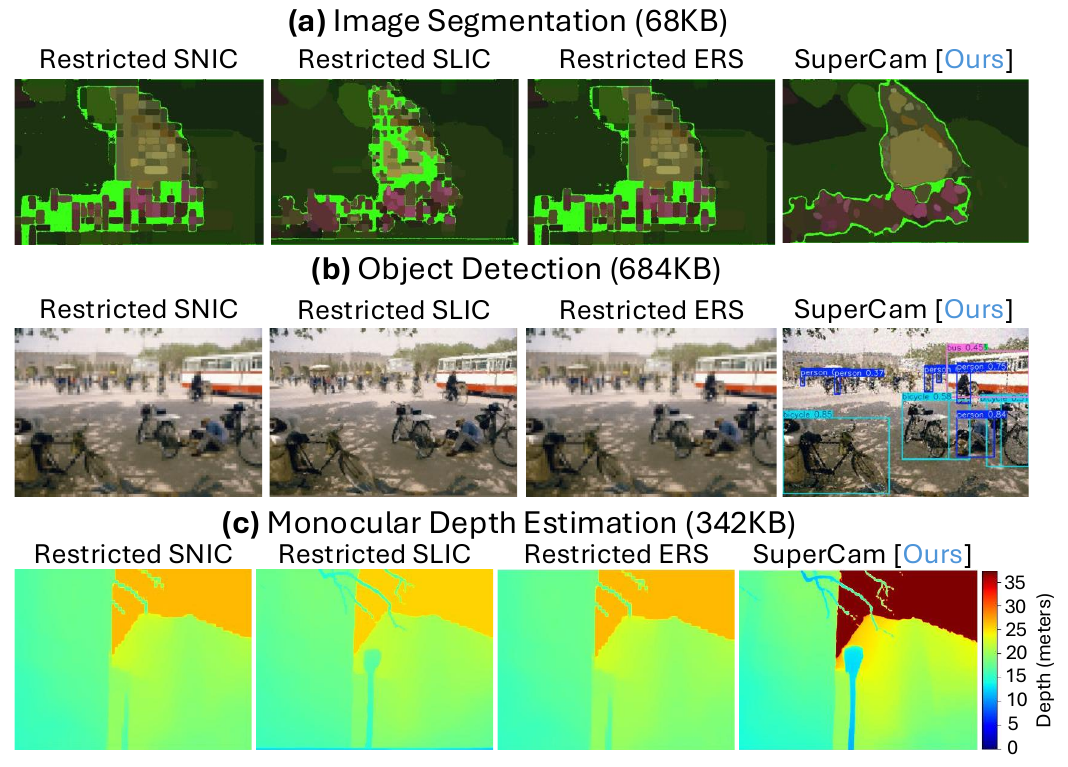}
    \caption{\textbf{Results of downstream computer vision applications for other non-learning based superpixel algorithms.} We show a snapshot of qualitative comparisons for SuperCam against memory restricted SNIC, ERS and SLIC. SuperCam performs better that all the other algorithms on all computer vision tasks.}
    \label{fig:Downstream_CV_Comparisons_Supplement}
\end{figure}

In the main paper we compared SuperCam results with memory restricted SNIC. Here we compare SuperCam with two recent learning based superpixel algorithms, SPAM \cite{spam}, LNSNet \cite{LNSNET}, and also two memory restricted non-learning based algorithms, SLIC \cite{SLIC} and ERS \cite{ERS}. Suppl. Fig. \ref{fig:Superpixel_UE_Comparisons_Supplement} shows quantitative results for the Under Segmentation Error for the two learning based superpixel algorithms and all four non-learning based methods at a range of memory levels. In Suppl. Fig. \ref{fig:Downstream_CV_Comparisons_Supplement} we also show a snapshot of the qualitative results for the downstream computer vision tasks of Image Segmentation with Segment Anything Model v2 \cite{SAM2}, Object Detection with YOLOv12 \cite{yolov12}, and Monocular Depth estimation with Depth Anything v2 \cite{DepthAnythingv2} using images from the BSD500 \cite{BSD500}, COCO \cite{COCO} and DIODE \cite{DIODE} datasets.

\clearpage\newpage
\section{Superpixel Evaluation}
\label{sec:Superpixel_Evaluation_Supplementary}

In this section we provide all the formulae for various metrics used to evaluate the superpixel segmentation algorithms.

The Under Segmentation Error is defined as:

\begin{equation}
    UE_{NP}(G,S) = \frac{1}{N} \sum_{G_i} \sum_{S_j \cap G_i \neq \varnothing} \min\{ \mid S_j \cap G_i \mid , \mid S_j - G_i \mid \}
\end{equation}

\noindent where $S_j$ where $1 \leq j \leq K$ and $G_i$ where $1 \leq i \leq K$ are the corresponding partitions of the same image and $N$ is the total number of pixels in the image. We use the implementation of the Under Segmentation Error given in \cite{superpixel-survey1}.

We also show superpixel algorithm performance using a precision vs recall plot. Precision is defined as:

\begin{equation}
    Pre(S,G) = \frac{TP}{TP + FP}
\end{equation}

\noindent where TP is the number of true positives and FP is the number of false positives. TP and FP are calculated as:

\begin{equation}
    TP(S,G) = \sum_{i=1}^{N} \mathds{1}_{j \epsilon \mathcal{N}(i,\epsilon)}(S_j, G_i)
\end{equation}

\begin{equation}
    FP(S,G) = \sum_{i=1}^{N} \left[ 1 -  \mathds{1}_{j \epsilon \mathcal{N}(i,\epsilon)}(S_j, G_i) \right]
\end{equation}

\noindent where $\mathcal{N}$ represents a $\epsilon \times \epsilon$ boundary around the pixel position $i$. $S_j$ where $1 \leq j \leq K$ and $G_i$ where $1 \leq i \leq K$ are the corresponding partitions of the same image and $N$ is the total number of pixels in the image. The function $\mathds{1}$ returns $1$ if a superpixel boundary overlaps with the ground truth boundary pixel within the neighborhood $\mathcal{N}$ and $0$ otherwise. $\epsilon$ is defined as $(2r + 1)$, where $r$ is $0.0025$ times the diagonal of the image rounded to the next integer. Recall can be calculated as:

\begin{equation}
    Rec(S,G) = \frac{TP}{TP + FN}
\end{equation}

\noindent where FN can be expressed as:

\begin{equation}
    FN(S,G) = (\sum_{i=1}^{N} G_i) - TP
\end{equation}

We found that applying a post-processing Gaussian blur to both the SuperCam and the SNIC superpixel images, improved performance for all the computer vision models we tested (Segment Anything Model 2 \cite{SAM2}, YOLOv12 \cite{yolov12} and DepthAnythingV2 \cite{DepthAnythingv2}). In order to calculate the optimal blur that has to be applied for the SuperCam images, we did both a theoretical and an empirical analysis. In the next section we show the explanation of the experiments we carried out.

\clearpage \newpage
\section{Deriving the Optimal Blur Kernel}
\label{sec:Blur_Kernel_Derivation_Supplement}

\begin{figure*} [!tb]
    \centering
    \includegraphics[width=\textwidth]{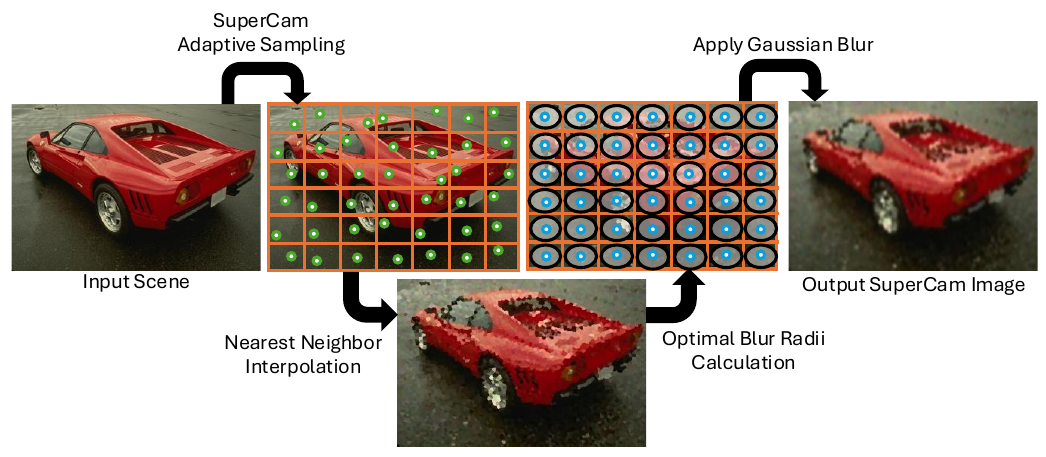}
    \caption{\textbf{SuperCam Algorithm.} We show a diagram of the proposed SuperCam algorithm. The pixels are first adaptively sampled to create an initial superpixel image that has holes. The holes are filled using nearest neighbor interpolation and a Gaussian Blur is applied to it to generate the final SuperCam image. The circles with the green border are the pixels that are exposed adaptively. The ellipses are the calculated optimal 2D Gaussian blur kernels for the pixel locations marked with blue borders.}
    \vspace{-0.1in}
    \label{fig:opt_blur_kernel}
\end{figure*}

We apply a post-processing Gaussian blur on the raw superpixel image produced by SuperCam. This improves the performance of all applications (image segmentation, object detection and monocular depth estimation) for both the SuperCam and SNIC images. We provide the theoretical and empirical derivation of the optimal blur for SuperCam. It is not possible to derive an optimal blur kernel for SNIC due to the absence of either a lower or upper bound on the superpixel size. However, empirically we found that the same size blur kernel derived for SuperCam also improves performance for SNIC superpixel images.

\subsection{Theoretical Derivation of the Optimal Blur Kernel for SuperCam}

Suppl. Fig. ~\ref{fig:opt_blur_kernel} shows a pictorial representation of the SuperCam algorithm. As the proposed SuperCam divides the image into rectangles of equal size and seeds each superpixel by a randomly chosen location within the rectangle, we know that the seeded location cannot be outside its corresponding grid. The intuition behind applying the Gaussian blur kernel is that doing so smooths the transition across superpixels. As each superpixel in a SegCam image will be within the initial grid of uniformly shaped rectangles, we apply a 2D Gaussian blur kernel with blur radii in each direction (width and height), equal to half of the width and height of the initial superpixel grid. The conversion from the blur radius to the standard deviation ($\sigma$) is done as follows:

\

\noindent A 1D Gaussian blur kernel used can be represented as: 

\begin{equation}
    g(x)=255 e^{-\frac{1}{2\sigma^{2}}x^{2}}
    \label{eq:gaussian_blur_eqn}
\end{equation}

\noindent where $\sigma$ is the standard deviation of the blur kernel applied and $x$ is the distance from origin. For a digital 1D signal the lowest value that is non-zero is 1, and we want this to happen exactly when we reach a distance of $r$, which is the known blur radius for that kernel. As the radius has to be non-inclusive, we substitute $r+1$ instead of $x$ in eqn \ref{eq:gaussian_blur_eqn}. Doing this we get:

\begin{equation}
    1 = 255 e^{-\frac{1}{2\sigma^{2}}(r+1)^{2}}
\end{equation}

\noindent Now rearranging and taking log on both sides, we end up with the relation:

\begin{equation}
    r = \sigma \sqrt{2 \log_{e}255} - 1
\end{equation}

This is the relation between the blur radius $r$ and the standard deviation of the blur kernel $\sigma$ for a 1D Gaussian blur operation. We use the fact that the 2D Gaussian blur kernel is separable to calculate the optimal blur kernel sigma for both the $x$ and $y$ dimensions of the image and apply two 1D Gaussian blur kernels. 

\subsection{Empirical Derivation of the Optimal Blur Kernel for SuperCam}
In order to confirm our theoretical calculation for the optimal blur radii, we conducted experiments testing image segmentation using several different multiples of our theoretical value. The resulting error metrics confirmed that the thoretically derived optimal blur kernel produced the lowest error.

\clearpage
\newpage
\section{Image Segmentation}
\label{sec:Image_Seg_Supplementary}

For validating the image segmentation results, we run Segment Anything Model V2 \cite{SAM2} on the BSD500 \cite{BSD500}, NYUV2 \cite{NYUV2} and SBD \cite{SBD} datasets and compare them using the mIOU Error metric. In this section we go through the definitions of the error metric used and also provide supplementary results to those show in the paper.
The mIOU Error is defined as:

\begin{equation}
    mIOU Error = \frac{1}{N} \sum_{i=1}^{N} 1 - \frac{S_i \cap G(S_i)}{S_i \cup G(S_i)} 
\end{equation}

where $i$ is the id of the segment in the segmented image, $N$ refers to the total number of segments in the segmented image, $S_i$ represents the segment with the id $i$ and $G(S_i)$ returns the ground truth segment that has maximum overlap with the segment $S_i$.

\begin{figure}[!h]
    \centering
    \includegraphics[width=0.5\linewidth]{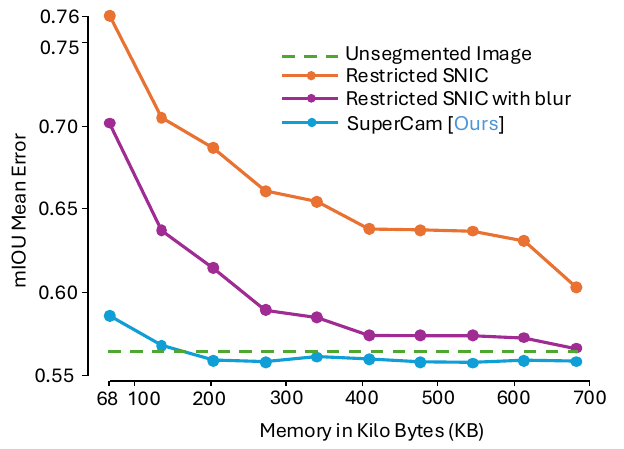}
    \caption{
    \textbf{Quantitative evaluation of image segmentation.} We compare the quality of image segmentation results produced by the SegmentAnythingV2 model using the mIOU metric on publicly available SBD dataset.
    Observe that our proposed SuperCam method achieves a lower mIOU mean error than SNIC.
    The error approaches that of an unsegmented image as the number of superpixels is increased.    
    }
    
    \label{fig:img_seg_eval_supplement}
\end{figure}

\begin{figure*}[!ht]
    \centering
    \includegraphics[width=0.97\linewidth]{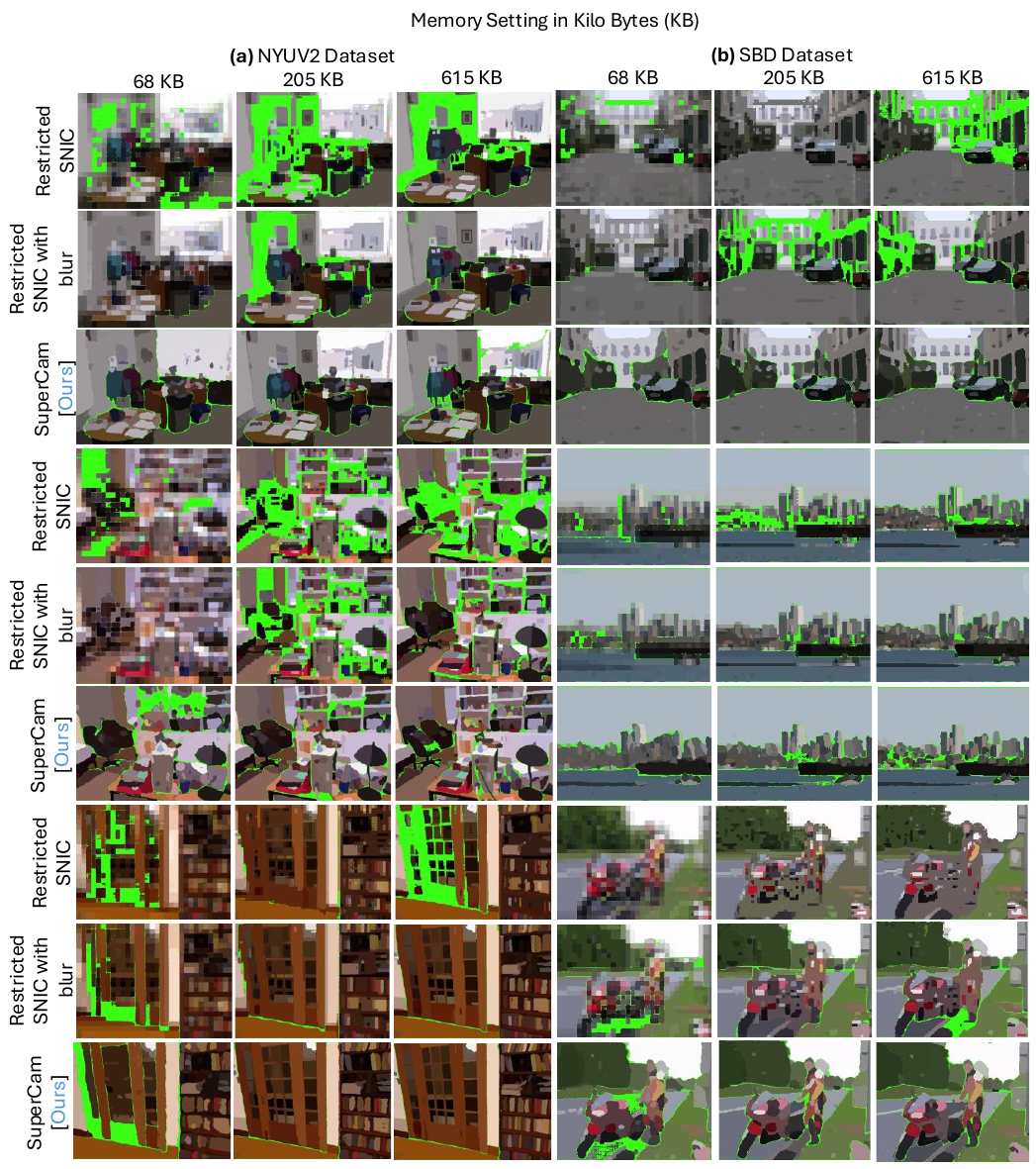}
    \caption{\textbf{Comparison of qualitative Image Segmentation results for memory-restricted SNIC and SuperCam.} Image segmentation results produced using the SAMV2 model on superpixel images for SuperCam, raw SNIC, and SNIC with the optimal blur kernel applied. Images are drawn from the \textbf{(a)} NYUV2 and \textbf{(b)} SBD datasets for different memory settings in kilobytes (KB). SuperCam results look visually better and converge to the ground truth as we increase the memory used.}
    \label{fig:Image_Segmentation_Supplement}
\end{figure*}

We show the mIOU Error plots for both the BSD500 \cite{BSD500} and the \cite{NYUV2} datasets in the paper. Suppl. Fig. \ref{fig:img_seg_eval_supplement} shows the error plot on the SBD \cite{SBD} dataset. The trends are similar to what we see in the other two datasets shown in the paper. SuperCam performs consistently better than SNIC and converges to the unsegmented image error as we increase the memory used. We also show more comparisions of SuperCam and SNIC in Suppl. Fig. ~\ref{fig:Image_Segmentation_Supplement}.

\clearpage\newpage
\section{Object Detection}
\label{sec:Object_Det_Supplementary}

We evaluate the object detection performance of YOLOv12 \cite{yolov12} on the COCO \cite{COCO} dataset with SuperCam and SNIC using the $mAP(50-95)$ score. This section provides background on the metric definition and additional qualitative results for object detection. The $mAP(50-95)$ score refers to the average of the mean average precision across a range of Intersection over Union (IOU) thresholds in increments of 0.05, starting from 0.50 (which denotes at least a 50\% overlap in the detected bounding boxes) to 95\% (which denotes at least a 95\% overlap in the detected bounding boxes).

\

The Intersection Over Union (IOU) is defined as:
\begin{equation}
    IOU = \frac{area \: of \: overlap}{area \: of \: union}
\end{equation}

\begin{figure*}[!ht]
    \centering
    \includegraphics[width=0.65\linewidth]{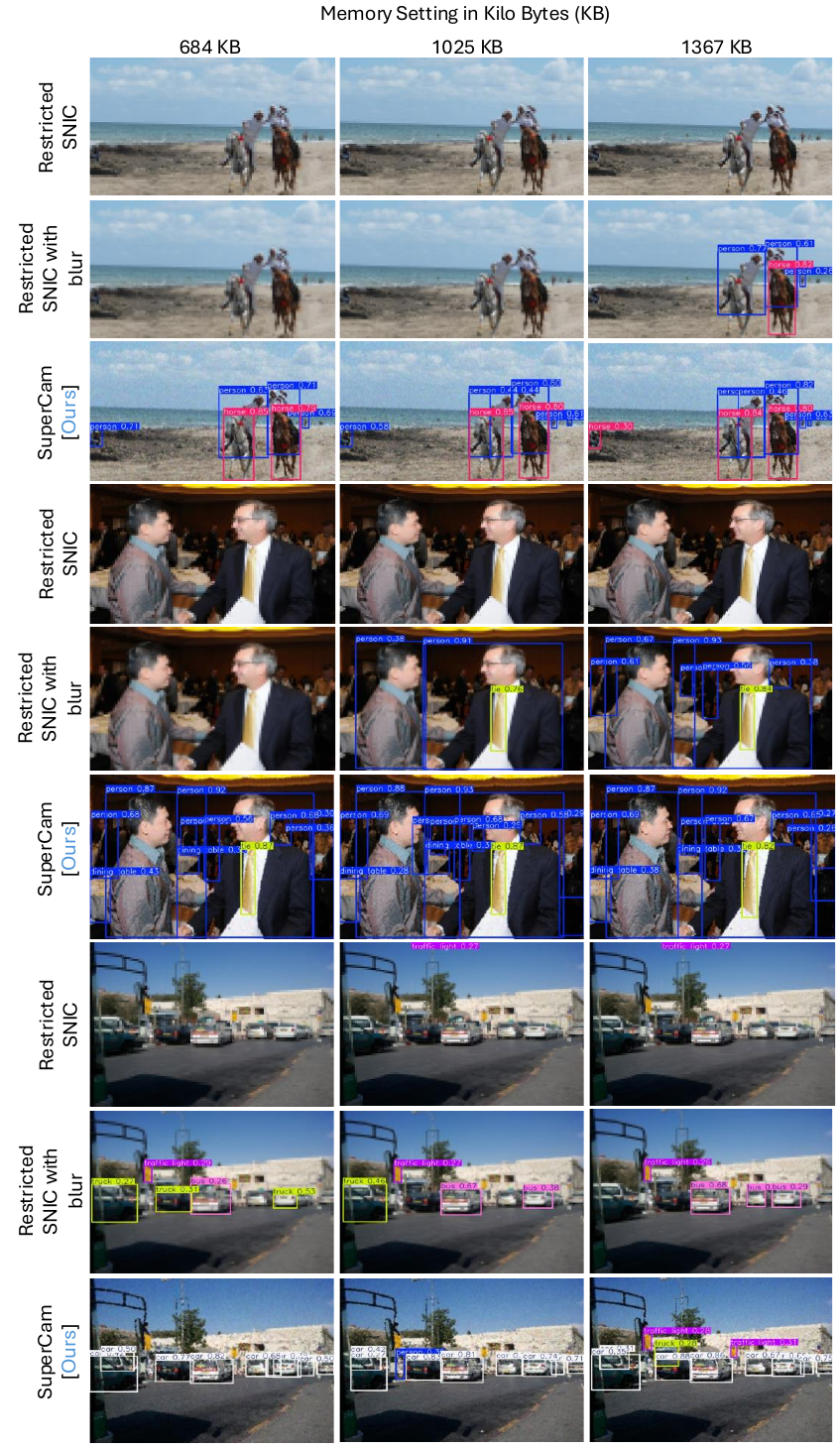}
    \caption{\textbf{Comparison of memory-constrained SNIC and SuperCam superpixel results for Object Detection.} Results on the YOLOv12 model applied to superpixel images for SuperCam, raw SNIC, and SNIC with the optimal blur kernel applied. Images are drawn from the COCO dataset, and each column shows a different memory setting in kilobytes (KB). SuperCam results look visually better and converge to the ground truth as we increase the memory used.}
    \label{fig:Object_Detection_Supplement}
\end{figure*}

Precision is defined as:

\begin{equation}
    Pre(S,G) = \frac{TP}{TP + FP}
\end{equation}

\noindent where TP is the number of true positives and FP is the number of false positives. Now $mAP50$ can be denoted as:

\begin{equation}
    mAP50 = \frac{1}{NC}\sum_{i=1}^{N}\sum_{j=1}^{C}Precision(D_{ij} * \mathds{1}(IOU(D_{ij})))
\end{equation}

\noindent where $N$ is the number of images in the dataset, $C$ is the number of categories in the dataset. $D_{ij}$ refers to the detection of category $j$ in the image $i$ and $\mathds{1}$ denotes a function that returns $1$ if $IOU(D_{ij}) \geq 50$, $0$ otherwise. Suppl. Fig. \ref{fig:Object_Detection_Supplement} shows additional object detection results for the validation partition of the COCO \cite{COCO} dataset using the YOLOv12 \cite{yolov12} model.

\clearpage

\clearpage\newpage
\section{Monocular Depth Estimation}
\label{sec:Mono_Depth_Supplementary}

We evaluate the performance of the DepthAnythingV2 \cite{DepthAnythingv2} model on the KITTI \cite{KITTI}, DIODE \cite{DIODE}, NYUV2 \cite{NYUV2} and the Sintel \cite{Sintel} dataset images generated by SuperCam and SNIC. Here we present the error metrics used to evaluate the estimated depth, show additional visual results, and discuss a few failure cases.

We use Absolute Relative Error and Threshold Accuracy to compare the results. Absolute Relative Error (AbsRel) measures how much the predict depth deviates from the ground truth in terms of percentages. Threshold Accuracy ($\delta_{1}$) measures the percentage of pixels that differ by no more than 25\%.

The predicted depth is compared with the ground truth depth available in the KITTI \cite{KITTI}, DIODE \cite{DIODE}, NYUV2 \cite{NYUV2} and the Sintel \cite{Sintel} datasets. The error metrics are defined as follows:

\begin{equation}
    AbsRel = \frac{1}{M}\sum_{j=1}^{M} \left( \frac{1}{N_j} \sum_{i=1}^{N_j} \frac{ \left| d_{ij} - \hat{d}_{ij} \right|}{d_{ij}} \right)
\end{equation}
\begin{equation}
    \delta_{1} = \frac{1}{M}\sum_{j=1}^{M}\left( \frac{1}{N_j} \sum_{j=1}^{N_j} \mathds{1}\left( \max\left( \frac{d_{ij}}{\hat{d}_{ij}}, \frac{\hat{d}_{ij}}{d_{ij}} \right) < 1.25 \right) \right)
\end{equation}

\noindent where $M$ denotes the number of images in the dataset, $N_j$ denotes the number of pixels in the image $j$, $d_{ij}$ denotes the ground truth depth at pixel $i$ and image $j$ and $\mathds{1}$ denotes the indicator function.

\begin{figure}[!bth]
    \centering
    \includegraphics[width=0.9\linewidth]{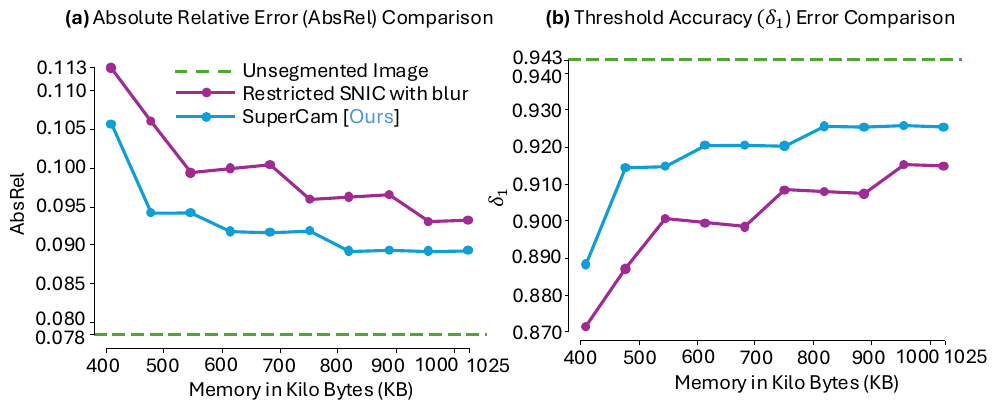}
    \caption{\textbf{Quantitative evaluation of monocular depth estimation on the KITTI dataset.} We compare the \textbf{(a)} Absolute Relative Error (AbsRel) and \textbf{(b)} Threshold Accuracy ($\delta_{1}$) metrics for depth estimates produced by the DepthAnythingV2 model on the KITTI dataset. Observe that SuperCam error metrics are better when compared to restricted SNIC across all memory settings.}
    \label{fig:mono_depth_eval_kitti_supplementary}
\end{figure}

\begin{figure}[!bth]
    \centering
    \includegraphics[width=0.9\linewidth]{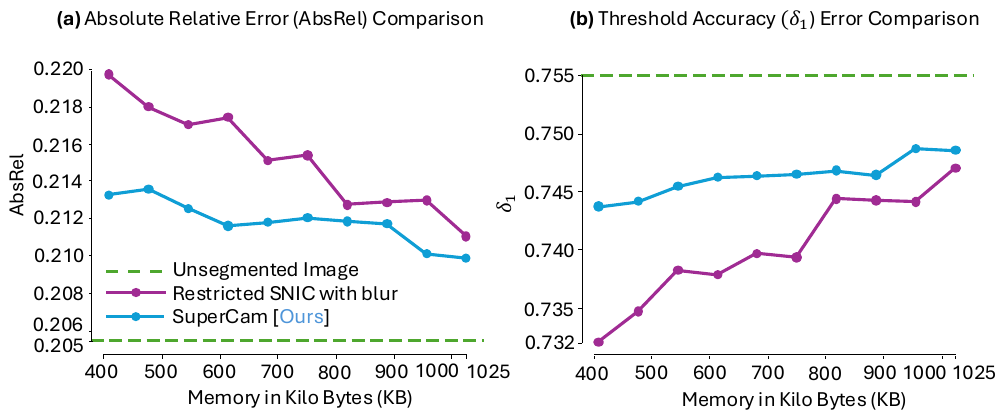}
    \caption{\textbf{Quantitative evaluation of monocular depth estimation on the DIODE dataset.} We compare the \textbf{(a)} Absolute Relative Error (AbsRel) and \textbf{(b)} Threshold Accuracy ($\delta_{1}$) error metrics for depth estimates produced by the DepthAnythingV2 model on the DIODE dataset. Observe that SuperCam error metrics are better when compared to restricted SNIC using the same amount of memory.}
    \label{fig:mono_depth_eval_diode_supplementary}
\end{figure}

\begin{figure}
    \centering
    \includegraphics[width=0.9\linewidth]{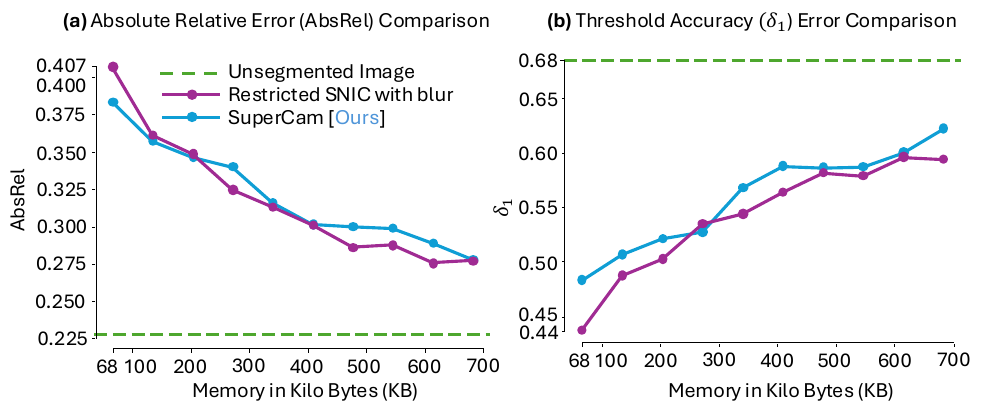}
    \caption{\textbf{Quantitative evaluation of mean absolute error for monocular depth estimation on the Sintel dataset.} We compare the \textbf{(a)} mean Absolute Relative Error (AbsRel) and \textbf{(b)} Threshold Accuracy ($\delta_{1}$) error metrics for depth estimates produced by the DepthAnythingV2 model on the Sintel dataset. AbsRel and Threshold Accuracy values are comparable to Restricted SNIC for all memory settings.}
    \label{fig:mono_depth_eval_sintel_supplementary}
\end{figure}

\begin{figure}
    \centering
    \includegraphics[width=0.9\linewidth]{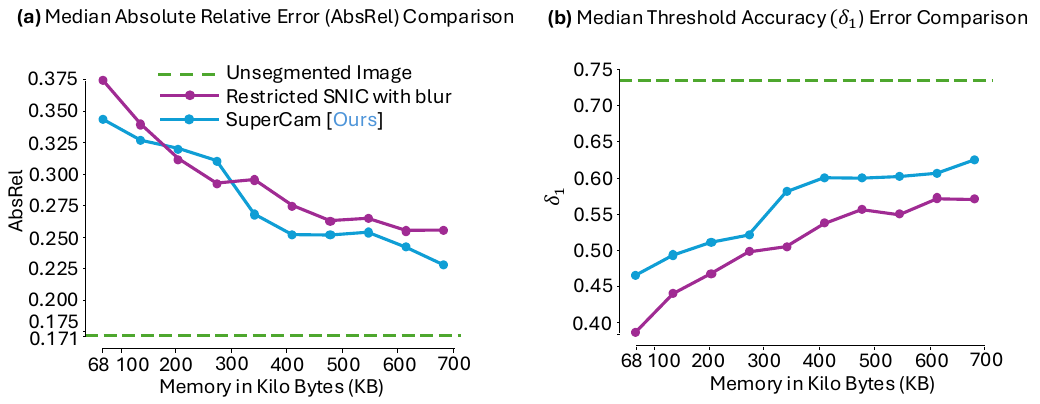}
    \caption{\textbf{Quantitative evaluation of median absolute error for monocular depth estimation on the Sintel dataset.} We compare the median \textbf{(a)} Absolute Relative Error (AbsRel) and \textbf{(b)} Threshold Accuracy ($\delta_{1}$) error metrics for depth estimates produced by the DepthAnythingV2 model on the Sintel dataset. AbsRel are comparable to or better than Restricted SNIC for all memory settings. Average Threshold Accuracy values are higher than Restricted SNIC across memory settings, although subsequent figures show that the underlying error distributions are comparable.}
    \label{fig:mono_depth_eval_sintel_median_supplementary}
\end{figure}

Suppl. Figs. \ref{fig:mono_depth_eval_kitti_supplementary} , \ref{fig:mono_depth_eval_diode_supplementary}, \ref{fig:mono_depth_eval_sintel_supplementary} show the $AbsRel$ and $\delta_{1}$ error metrics for the monocular depth estimates for the DepthAnythingV2 \cite{DepthAnythingv2} model on the KITTI \cite{KITTI}, DIODE \cite{DIODE} and Sintel \cite{Sintel} datasets. SuperCam images give better $AbsRel$ and $\delta_1$ errors than SNIC for the KITTI \cite{KITTI}, DIODE \cite{DIODE} and NYUV2 \cite{NYUV2} datasets. For the Sintel \cite{Sintel} dataset, SuperCam gives better $\delta_1$ errors and similar $AbsRel$ errors. The mean $AbsRel$ error values plotted for SuperCam are high due to outlier error values produced by a few images. Some SuperCam images have a few pixels that have very high or very low estimated depth values when compared to the ground truth. 
Fig. \ref{fig:mono_depth_eval_sintel_median_supplementary} shows the median error plots for AbsRel and $\delta_{1}$ for the Sintel \cite{Sintel} dataset. It can be seen that apart from two data points in the AbsRel plot (for memory settings 205KB and 273KB) all other data points show better results for SuperCam. 

\begin{figure*}[!p]
    \centering
    \vspace{-30pt}
    \includegraphics[width=0.68\linewidth]{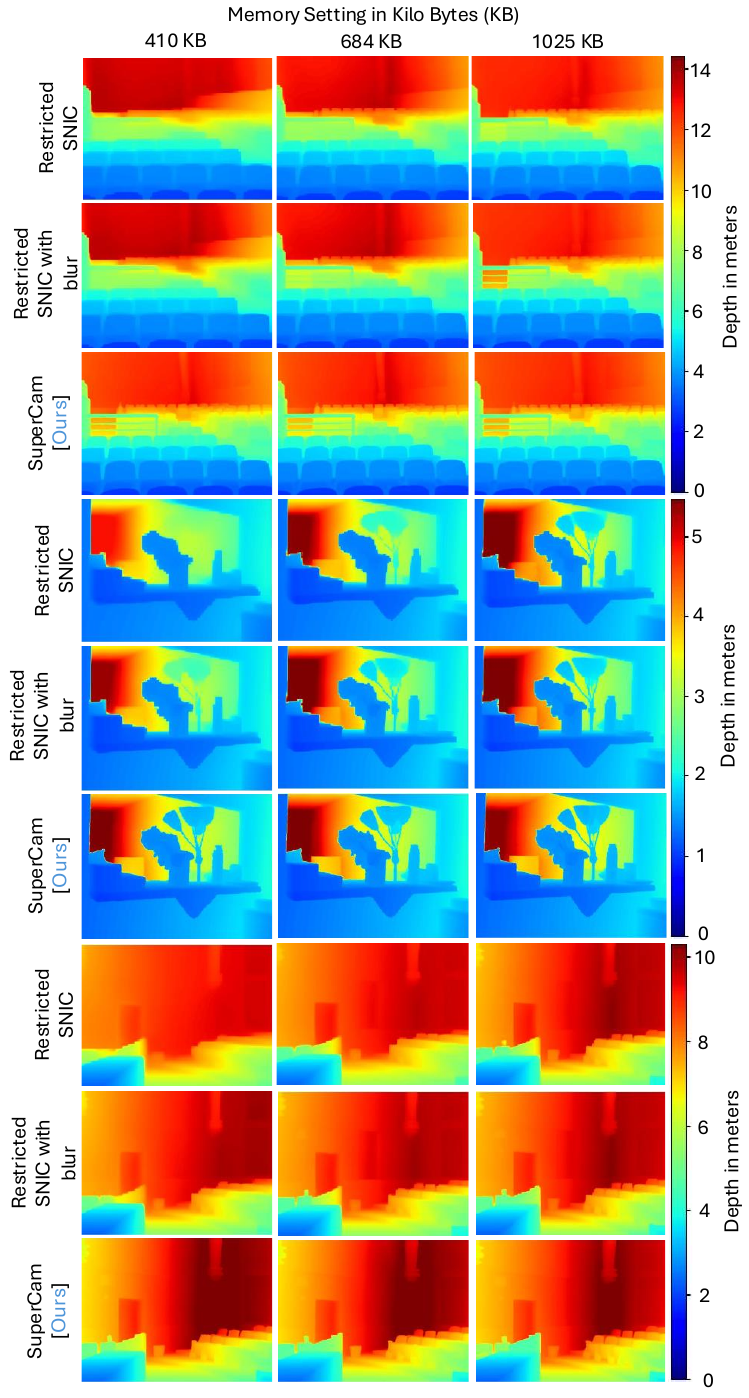}
    \caption{\textbf{Comparison of Monocular Depth Estimation results for the DIODE dataset.} Object Detection results on the DepthAnythingV2 model applied to the superpixel images for SuperCam, Restricted SNIC, and SNIC with the optimal blur kernel applied. Images are drawn from the DIODE dataset and columns show different memory settings in kilobytes (KB). SuperCam results look visually better, particularly for lower memory settings.}
    \label{fig:Mono_Depth_Supplement1}
\end{figure*}

\begin{figure*}[!p]
    \centering
    \includegraphics[width=0.90\linewidth]{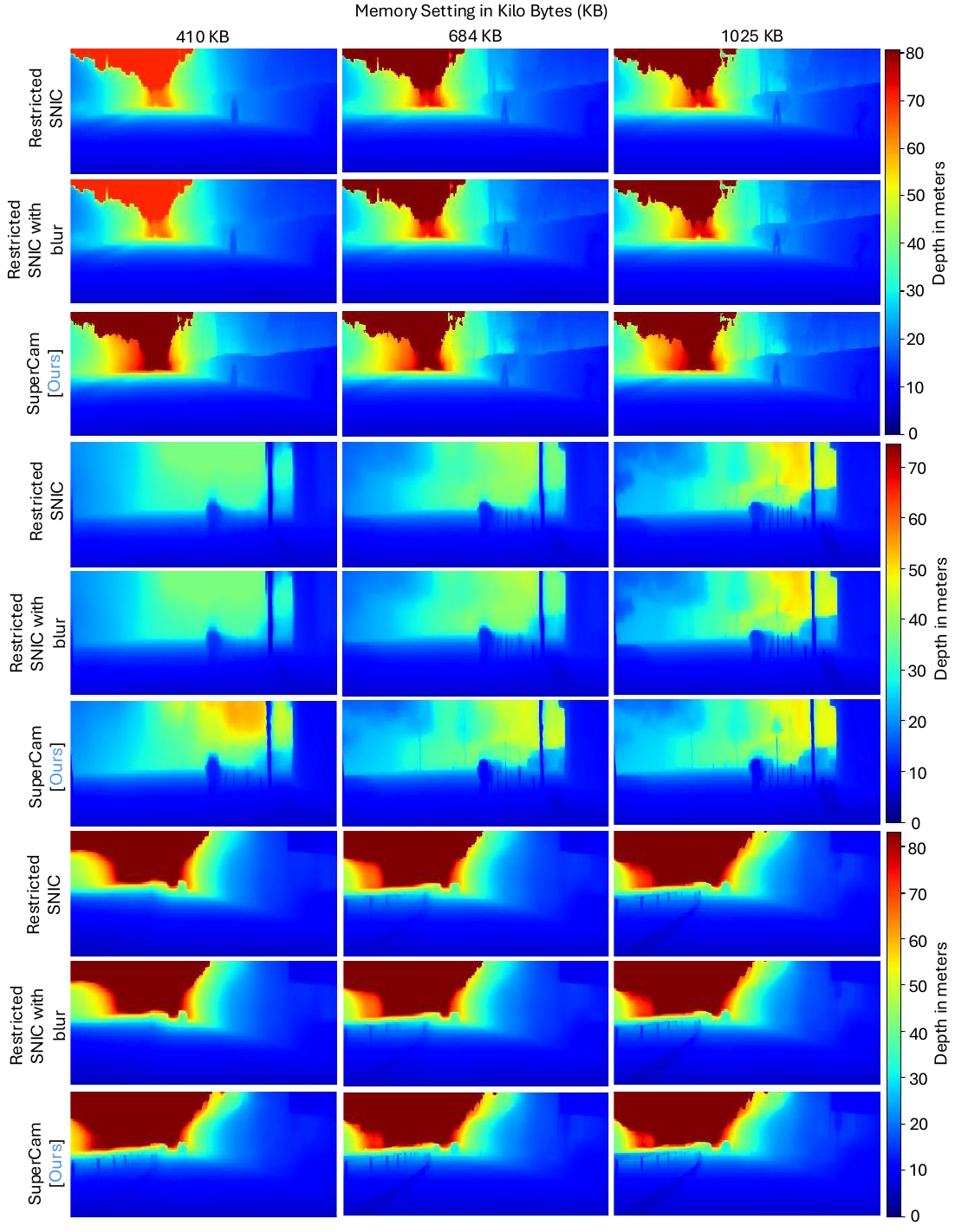}
    \caption{\textbf{Comparison of Monocular Depth Estimation results for the KITTI dataset.} Object Detection results on the DepthAnythingV2 model applied to superpixel images for SuperCam, Restricted SNIC, and SNIC with the optimal blur kernel applied. Images are drawn from the KITTI dataset, and columns show different memory settings in kilobytes (KB). SuperCam results look slightly better visually. The lack of significant difference visually between SuperCam and SNIC can be explained by the fact that the KITTI dataset is very sparse and this affects the conversion of relative depth to absolute depth as there are only a few valid data points to do the conversion.}
    \label{fig:Mono_Depth_Supplement2}
\end{figure*}

\begin{figure*}[!ht]
    \centering
    \includegraphics[width=0.90\linewidth]{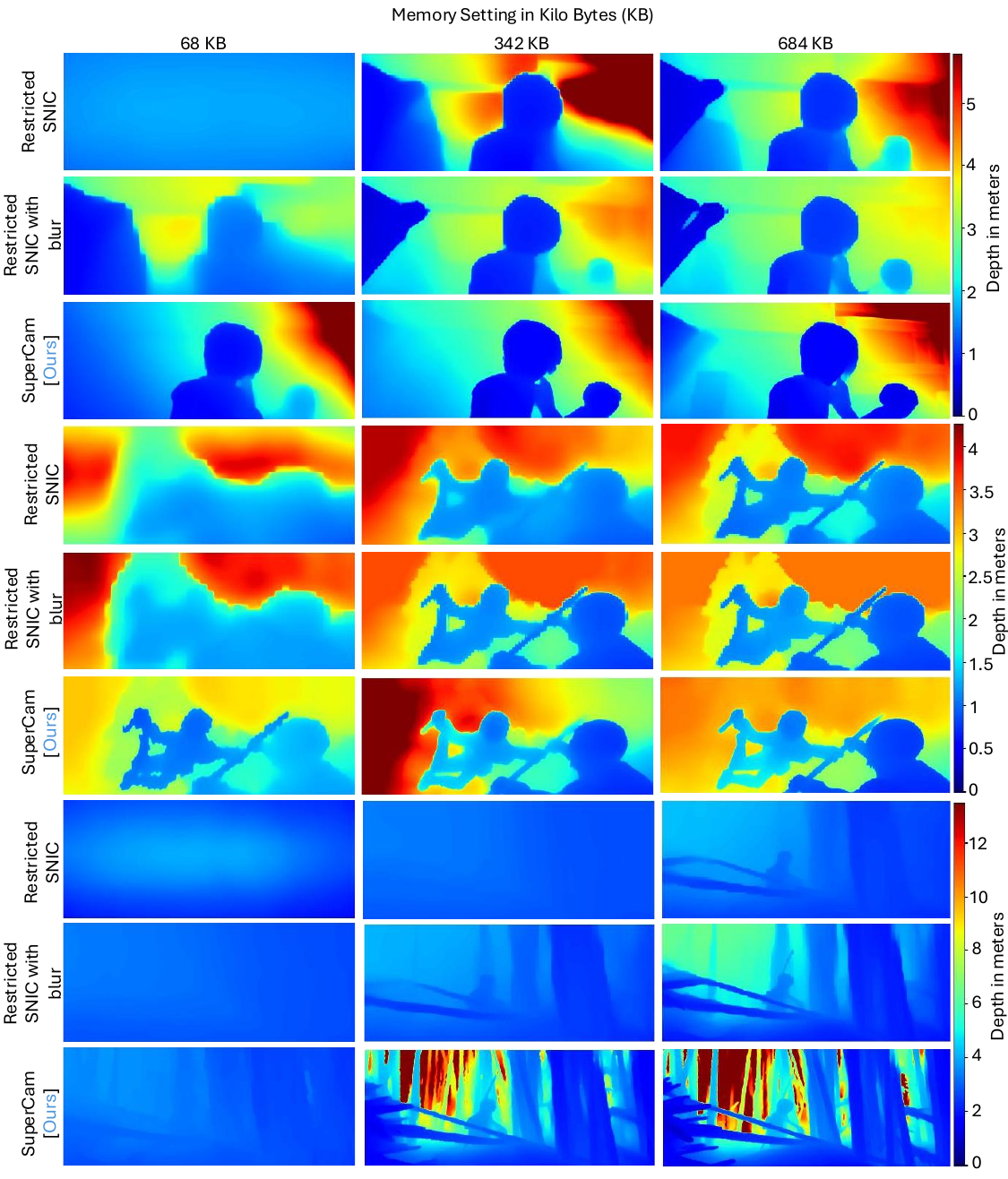}
    \caption{\textbf{Comparison of Monocular Depth Estimation results for the Sintel dataset.} Object Detection results on the DepthAnythingV2 model on the superpixel images for SuperCam, Restricted SNIC, and SNIC with the optimal blur kernel applied. Images are drawn from the Sintel dataset and columns show different memory settings in kilobytes (KB). SuperCam results look visually better.}
    \label{fig:Mono_Depth_Supplement3}
\end{figure*}

Suppl. Figs. \ref{fig:Mono_Depth_Supplement1}, \ref{fig:Mono_Depth_Supplement2} and \ref{fig:Mono_Depth_Supplement3} show depth estimates given by the DepthAnthingV2 \cite{DepthAnythingv2} model on the DIODE \cite{DIODE}, KITTI \cite{KITTI} and Sintel \cite{Sintel} datasets.

\clearpage\newpage
\section{Real Hardware Results}
So far we have synthetically simulated the images that SuperCam would capture from a normal RGB image. We simulate the photon streams corresponding to every pixel and generate the SuperCam image from the simulated data. Next we show how a SuperCam would perform when real-world captures are used.

\begin{figure*}[!ht]
    \centering
    \includegraphics[width=0.90\linewidth]{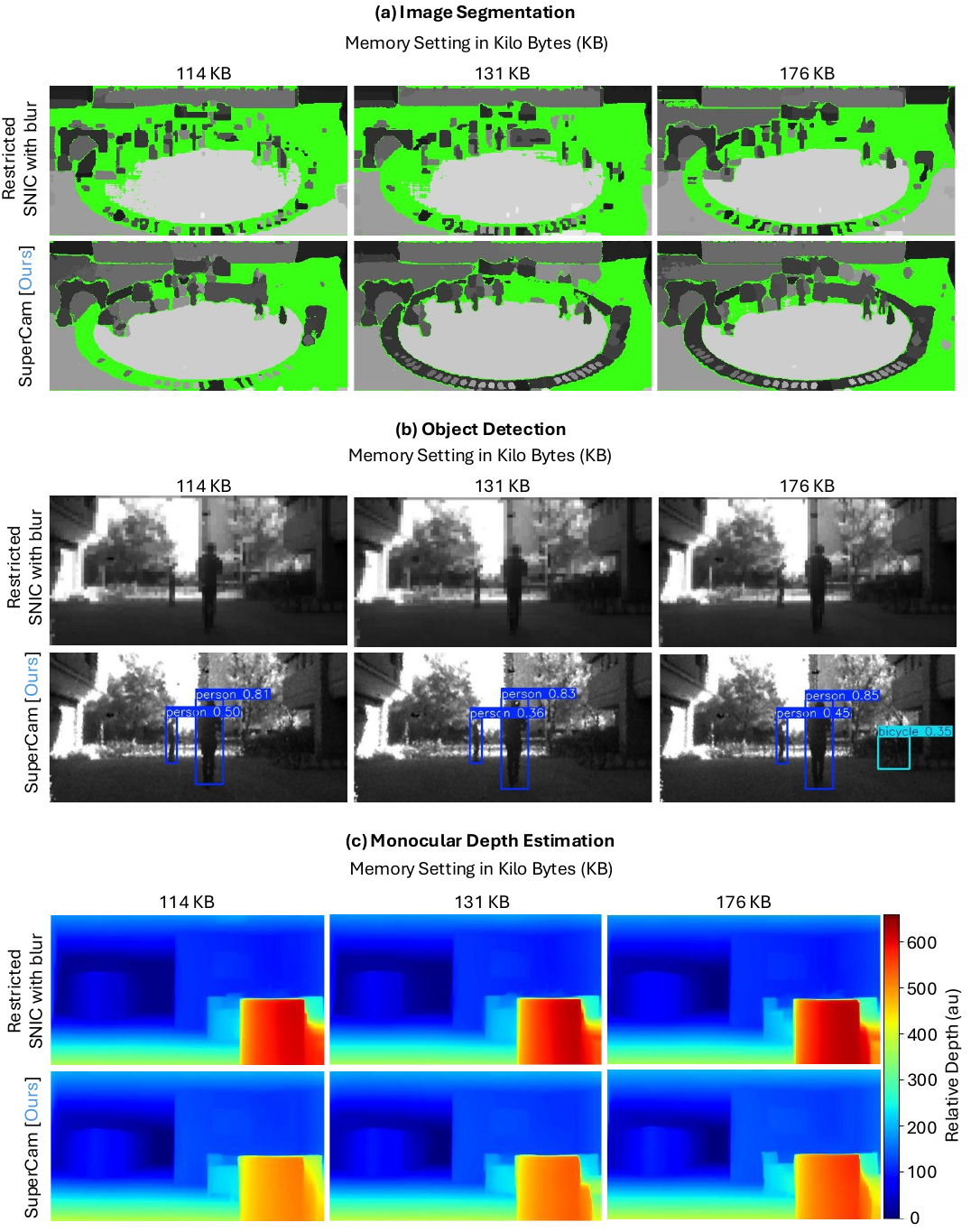}
    \caption{\textbf{Qualitative comparison of experimental results.} We show experimental results of SuperCam on real world SPAD data from the SWISS SPAD sensor used in the Burst-Vision work. \textbf{(a)} Image Segmentation results using SAM2 on experimental SPAD data collected from the SWISS SPAD sensor. The florescent green color refers to regions of the image that were not given any mask by SAM2 model. \textbf{(b)} Object Detection results using the YOLOv12 model on experimental SPAD data collected from the SWISS SPAD sensor. \textbf{(c)} Monocular Depth Estimation results using the DepthAnythingV2 model on experimental SPAD data collected from the SWISS SPAD sensor. Observe that results the SuperCam results are visually better than the SNIC result that uses the same amount of memory. The results improve with increase in the amount of memory used.}
    \label{fig:Real_World_Results_Supplement}
\end{figure*}

Lastly, we provide results obtained on a publicly available real-world dataset, \cite{burst_vision} captured using a SwissSPAD sensor \cite{ulku2017512}. We compare SuperCam results with memory restricted SNIC on the captured sequences. Suppl. Fig.~\ref{fig:Real_World_Results_Supplement} shows the results obtained using the captured sequences. It can be observed that SuperCam results are visually better when compared to SNIC for image segmentation and object detection. For monocular depth estimation, we show only relative depth as we do not have ground truth for the captured scenes. The depthmaps for SNIC are pixelated and blocky when compared to the SuperCam results.

To summarize all the real hardware results and to show that they have temporal consistency, we also provide a short video sequence (named ``supercam\_demo\_video.mp4") of all three computer vision tasks; image segmentation, object detection and monocular depth estimation on a captured sequence that is part of the publicly available real-world dataset \cite{burst_vision}. The video sequence is captured in high dynamic range lighting conditions, has motion-blur due to movement of the camera, subjects and objects in the scene and are noisy. Nevertheless, SuperCam performance is comparable to the unsegmented image and is visually better than the memory-restricted SNIC results. For image segmentation, the area of frame that is unsegmented is less for SuperCam when compared to SNIC. The segmented regions are also more consistent and similar to the unsegmented image results. For object detection, SuperCam gets more detections than SNIC and misses only the relatively small objects among the objects detected in the unsegmented video. Compared to this the SNIC video performs poorly. For monocular depth estimation, we show the relative depth for the captured scene. The SNIC results are pixelated, especially along the boundaries of objects, when compared to SuperCam. Note that there is some flickering in the image segmentation results for all methods including the unsegmented video. This is caused by the photon noise present in the reconstructed image from the photon cube.

\end{document}